\documentclass[preprint,12pt,authoryear]{elsarticle}

\usepackage{setspace}
\usepackage{geometry} 
\usepackage{makecell}
\usepackage{booktabs}
\usepackage{amsmath}
\usepackage{amssymb}
\usepackage{epstopdf}
\usepackage[labelsep=period]{caption}  
\usepackage[british]{babel} 
\usepackage[hang]{footmisc}
 
\usepackage{hyperref}
\hypersetup{
    colorlinks=true,
    linkcolor=blue,
    filecolor=magenta,      
    urlcolor=cyan,
    pdftitle={Overleaf Example},
    pdfpagemode=FullScreen,
    }

\usepackage{url}
\usepackage{comment}
\usepackage[noend]{algpseudocode} 
\usepackage{graphicx}
\usepackage{subcaption}
\usepackage{lineno}
\usepackage{etoolbox}
\AtBeginEnvironment{algorithmic}{\ttfamily}

\begin{document}
\begin{frontmatter}

\title{Inferring Height from Earth Embeddings: \\ 
First insights using Google AlphaEarth
}

\author{
 Alireza Hamoudzadeh\textsuperscript{* 1},  
Valeria Belloni\textsuperscript{1}, 
Roberta Ravanelli\textsuperscript{2} }

\address{
	\textsuperscript{1 }Geodesy and Geomatics Division, DICEA, Sapienza University of Rome, Rome, 00184, Italy\\
	\textsuperscript{2 }Geomatics Unit, Department of Geography, Faculty of Sciences, University of Liège, Liège, 4000, Belgium\\
    \textsuperscript{* }Corresponding author: \href{mailto:alireza.hamoudzadeh@uniroma1.it}{alireza.hamoudzadeh@uniroma1.it}
}

\begin{abstract}
This study investigates whether the geospatial and multimodal features encoded in \textit{Earth Embeddings} can effectively guide deep learning (DL) regression models for regional surface height mapping.
In particular, we focused on AlphaEarth Embeddings at 10 m spatial resolution and evaluated their capability to support terrain height inference using a high-quality Digital Surface Model (DSM) as reference. U-Net and U-Net++ architectures were thus employed as lightweight convolutional decoders to assess how well the geospatial information distilled in the embeddings can be translated into accurate surface height estimates. Both architectures achieved strong training performance (both with $R^2 = 0.97$), confirming that the embeddings encode informative and decodable height-related signals. 
On the test set, performance decreased due to distribution shifts in height frequency between training and testing areas. Nevertheless, U-Net++ shows better generalization ($R^2 = 0.84$, median difference = -2.62 m) compared with the standard U-Net ($R^2 = 0.78$, median difference = -7.22 m), suggesting enhanced robustness to distribution mismatch. While the testing RMSE (approximately 16 m for U-Net++) and residual bias highlight remaining challenges in generalization, strong correlations indicate that the embeddings capture transferable topographic patterns.
Overall, the results demonstrate the promising potential of AlphaEarth Embeddings to guide DL-based height mapping workflows, particularly when combined with spatially aware convolutional architectures, while emphasizing the need to address bias for improved regional transferability.

\end{abstract}

\end{frontmatter}

\sloppy

\section{Introduction}
\label{sec:Introduction}

Geospatial image analysis is a rapidly evolving field at the intersection of Remote Sensing (RS), Computer Vision (CV), and Artificial Intelligence (AI), supporting a wide range of applications, from environmental and climate monitoring to urban modeling and infrastructure assessment.
In recent years, the field has entered a big data era, driven by the global acquisition of Earth Observation (EO) data at unprecedented spatial, temporal, and spectral resolutions. 
Although this new generation of EO data offers considerable potential for transformative geospatial insights by providing regular, consistent spatial information on the Earth's surface, its applicability in supervised data-driven frameworks is constrained by the limited availability of high-quality labeled data. This scarcity arises from the substantial cost and effort required for in situ measurements and/or expert interpretation \citep{Zhu2026, brown2025AlphaEarthfoundationsembeddingfield}. 

In parallel, EO data are inherently heterogeneous, necessitating the retrieval, alignment, and integration of data from diverse sensing modalities, such as multispectral imagery, Synthetic Aperture Radar (SAR), and Light Detection And Ranging (LiDAR), as well as from multiple acquisition platforms, including drones, aircraft, and satellites operating at different spatial scales. These challenges are further amplified by significant domain variability arising from differences in sensors, imaging characteristics and environmental conditions.
Moreover, although imagery constitutes a central component of geospatial analysis, it represents only one element of an increasingly rich geospatial data ecosystem. Additional sources, such as geospatial time series (e.g., GNSS displacements and meteorological observations) and textual data associated with geospatial information (e.g., annotations, reports, and sensor metadata), can provide complementary insights. However, effectively integrating heterogeneous modalities into a coherent analytical framework remains difficult and is often underexplored.

The growing imbalance between the abundance of EO data and the scarcity of labels, together with the need to integrate heterogeneous data sources for downstream tasks, has driven substantial interest in semi-supervised and unsupervised Deep Learning (DL) approaches for geospatial analysis. 
A key outcome of these efforts is the development of Geospatial Foundation Models (GeoFMs), large-scale DL architectures, typically pre-trained in a self-supervised or weakly supervised manner on diverse EO data. GeoFMs are designed to generalize across tasks, sensors, and areas while reducing dependence on task-specific supervision \citep{simumba2025, Zhu2026}. 
In this context, GeoFMs have gained increasing attention for their ability to capture shared patterns across heterogeneous geospatial data by embedding them into a unified latent space.
While other large-scale multimodal DL systems produce standard embeddings, GeoFMs uniquely enable the generation of location-aware embeddings, commonly referred to as \textit{Earth Embeddings} or \textit{Geospatial Embeddings} (Section \ref{sec:EarthEmbeddings}).
Such representations encode geographic and multimodal information directly into compact, learnable vectors \citep{Klemmer-2025,Klemmer-2025b}. 
By providing normalized location-specific representations within a shared metric space, \textit{Earth Embeddings} enable scalable, transferable, and modality-agnostic geospatial analysis across a wide range of downstream tasks. 
Recent studies have demonstrated the potential of \textit{Earth Embeddings} for various downstream tasks, including tree species classification, parcel segmentation, biomass estimation, and canopy height regression (see Section \ref{sec:AlphaEarth_downstream}).  

The present study investigates the suitability of \textit{Earth Embeddings} for a different task, i.e., surface height mapping.  
Specifically, we investigated how the geospatial and multimodal features encoded in \textit{Earth Embeddings} can guide DL-based regression models to predict morphological information derived from high-quality regional Digital Surface Models (DSMs).
Our analysis focused on the AlphaEarth Embedding dataset, the \textit{Earth Embeddings} dataset generated from the GeoFM \textit{AlphaEarth Foundations} of Google \citep{brown2025AlphaEarthfoundationsembeddingfield} and made available through Google Earth Engine (GEE) \citep{GORELICK2017}. 
These embeddings are derived from Sentinel-1 and Sentinel-2 imagery, complemented by auxiliary data sources, including height-related products such as Global Ecosystem Dynamics Investigation (GEDI) \citep{GEDI02_Av002}, Copernicus DEM \citep{ESA_c5d3d65}, and Gravity Recovery and Climate Experiment (GRACE) \citep{GRACE_Freshwater_Trends_2019}. The inclusion of such auxiliary information is expected to enhance the capacity of the embeddings to represent height-related features.  
To the best of our knowledge, this paper represents the first evaluation of \textit{Earth Embeddings} for surface height mapping. 

The main contributions of this paper are as follows:

\begin{itemize}
    \item a structured and critical review of prominent \textit{Earth Embeddings} datasets and recent studies employing AlphaEarth Embeddings for geospatial downstream tasks;
    \item the first systematic investigation of the feasibility of regional surface height mapping using AlphaEarth Embeddings within a DL-based regression framework, exploring how embedding-derived geospatial representations can effectively support and inform terrain height inference from high-quality regional DSMs.
\end{itemize}

\section{Earth Embedding}\label{sec:EarthEmbeddings}

Formally, an embedding is a vector representation of data in a continuous $n$-dimensional space, where each vector corresponds to a set of normalized coordinates within that space. 
Embeddings are obtained by projecting the original high-dimensional data into a lower-dimensional, semantically meaningful space through a learned compression process. In this representation, semantically similar samples are mapped to nearby locations (i.e., coordinates) in the embedding $n$-dimensional space, thereby facilitating tasks such as similarity search and clustering. 

Similarly, \textit{Earth Embeddings} are vector representations of geospatial data associated with specific locations in space and time \citep{Klemmer-2025b}, where each location is encoded as a set of normalized coordinates in the considered continuous $n$-dimensional space for a given epoch. \textit{Earth Embeddings} can be interpreted as condensed representations of diverse geospatial information, distilling complementary patterns from multi-temporal, multimodal data sources into a unified, information-rich representation. 
These deep features summarize essential characteristics of geographic locations, facilitating the quantification of similarities and differences in space and time, and supporting robust generalization across regions, spatial and temporal scales, and modalities. 
Overall, \textit{Earth Embeddings} serve as a consistent and scalable interface between complex geospatial data and GeoAI models, enabling the integration of heterogeneous datasets across a wide range of applications. 
By providing a unified, information-rich representation, \textit{Earth Embeddings} have the potential to transform traditional geospatial mapping workflows, which often rely on large task-specific training datasets, computationally intensive models, and custom inference pipelines.

\textit{Earth Embeddings} can be generated by a variety of large-scale, multimodal DL systems, particularly GeoFMs, such as Prithvi \citep{Prithvi}, Copernicus-FM \citep{CopernicusFM}, Galileo \citep{Galileo}, SMARTIES \citep{Smarties}, and CROMA \citep{Croma}.  
GeoFMs are trained on massive volumes of geospatial data to learn transferable representations that can be fine-tuned for a wide range of downstream tasks, including image segmentation and classification \citep{Klemmer-2025b}. 
Consequently, \textit{Earth Embeddings} are a natural by-product of global-scale GeoFMs.

A key advantage of this paradigm is that embeddings can be computed once and distributed as compact descriptors, upon which lightweight, task-specific decoders can be trained.
This approach reduces computational costs and improves scalability, enabling efficient deployment across multiple applications. This shift from end-to-end task-specific GeoFM inference toward the reuse of precomputed \textit{Earth Embeddings} has led to the release of \textit{Earth Embeddings} as dedicated datasets generated by specifically designed GeoFMs. 
Notable examples include AlphaEarth \citep{brown2025AlphaEarthfoundationsembeddingfield} and TESSERA \citep{feng2025tessera}, which aim to provide global or future global coverage using largely open-source data, such as Sentinel-1 and Sentinel-2 imagery (Table \ref{tab1:Embedding}). 

\begin{table}[ht!]
\centering
\caption{Features of main Earth Embedding datasets.}

\begin{tabular}{lll}
\toprule
\textbf{Attribute} & \textbf{AlphaEarth} & \textbf{TESSERA} 
\\
\hline
\textbf{Spatial coverage} 
& Global 
& \makecell[l]{Global \\ (sparse)}\\ 
\hline

\textbf{Temporal coverage} 
& 2017-2024 
& \makecell[l]{2017-2024 \\ (not continuous)} 
\\
\hline

\textbf{Time granularity} 
& Annual 
& Annual 
\\
\hline

\textbf{Spatial resolution} 
& 10 m
& 10 m
\\
\hline

\textbf{Embedding dimension} 
& 64 bands
& 128 bands
\\
\hline

\textbf{Modality}
& Multi-modal 
& Multi-modal 
\\
\hline
\textbf{Training data}
& \makecell[l]{Sentinel-1 \\ Sentinel-2 \\ Landsat 8/9 \\ PALSAR-2 \\ 
ERA5-Land \\
GEDI \\ 
GRACE \\ 
Copernicus DEM \\
NLCD \\
Text sources}
& \makecell[l]{ Sentinel-1 \\ Sentinel-2}
\\
\hline

\textbf{Model input}
& Patch
& Pixel
\\

\hline

\textbf{Foundation Model} 
&  \makecell[l]{ AlphaEarth \\ Foundations}
&  Tessera
\\
\hline

\textbf{Open-source model} 
& No  
& Yes 
\\
\hline

\textbf{Open-source data} 
& Yes  
& Yes 
\\
\hline

\textbf{Release} 
& 2025
& 2025 
\\
\hline
\end{tabular}

\label{tab1:Embedding}
\label{tab1}
\end{table}

In addition, several other initiatives have released \textit{Earth Embedding} datasets, including MajorTOM \citep{czerkawski2024}, MOSAIKS \citep{mosaiks}, Clay \citep{Clay,Clay1}, S2Vec \citep{s2vec, Gramacki_SRAI_Towards_Standardization_2023}, Earth Index Embeddings \citep{EarthIndexEmbeddings2025},  Presto \citep{zvonkov2025,presto} and Embedded Seamless Data \citep{chen2026esd, chen2026esd2}.

The recent and rapid growth of \textit{Earth Embeddings} datasets has led to a highly fragmented ecosystem, characterized by substantial heterogeneity in data formats, spatial and temporal coverage, and the DL architectures used to generate them (Table \ref{tab1:Embedding}).
In addition, the absence of a standardized interface for accessing and evaluating the embedding products represents a significant limitation. 
To address these gaps, recent efforts have focused on developing tools and benchmarks for \textit{Earth Embeddings}. 
TerraTorch \citep{zadrozny2025, terratorch} recently introduced new functionality for embedding-centric workflows, while \cite{fang2026} implemented standardized data loaders within TorchGeo \citep{TorchGeo}, enabling consistent and efficient access to heterogeneous \textit{Earth Embedding} products through a unified Application Programming Interface (API).
Complementing these initiatives, NeuCo-Bench \citep{vinge2025} provided the first benchmark frameworks for assessing lossy neural compression and representation learning approaches in the EO domain. 
Together, these developments represent important steps toward improving interoperability, reproducibility, and systematic evaluation within the rapidly expanding \textit{Earth Embedding} ecosystem. 

\section{AlphaEarth Embeddings for downstream tasks}\label{sec:AlphaEarth_downstream}

By compressing correlated multisource geospatial data into compact vector representations, \textit{Earth Embeddings} accelerate experimentation and simplify GeoAI pipelines by enabling scalable workflows across different geospatial domains, including ecology, agriculture, urban studies, and climate science. 
Building on the growing adoption of \textit{Earth Embeddings} as compact, information-rich representations of geospatial data, AlphaEarth Embeddings have emerged as a widely used resource for downstream geospatial applications, thanks to the global coverage and availability through GEE.
This section reviews recent studies that investigate the use of AlphaEarth Embeddings in various downstream tasks. 

First introduced to the EO community in \cite{brown2025AlphaEarthfoundationsembeddingfield}, AlphaEarth Embeddings demonstrated that AlphaEarth Foundations GeoFM delivers strong performance on multiple EO applications, primarily land cover and land use classification, without task-specific retraining.
Subsequent studies have extended the evaluation of AlphaEarth Embeddings to a broader set of geospatial tasks, benchmarking their performance against conventional remote sensing features and models. The key findings are described below by application domain.

\paragraph{Land Cover and Land Use Classification}

\cite{Khan} evaluated AlphaEarth Embeddings for land cover and land use mapping over a 19.3 km$^{2}$ region in Pakistan. Using both pixel-based and object-based Random Forest (RF) pipelines, they reported that AlphaEarth Embeddings consistently outperformed traditional workflows based on spectral indices, improving overall accuracy by up to $5\%$ and Cohen’s Kappa by approximately $3\%$.  

\paragraph{Environmental Monitoring and Air Quality Modeling}

\cite{Alvarez2025} used AlphaEarth Embeddings as the sole predictors in machine learning models to estimate urban air pollutants, including NO$_{2}$, SO$_{2}$, PM$_{2.5}$, and CO. The analyses focused on 60 points for each pollutant in Quito city (Ecuador), where data from stations were available.
Support Vector Regression (SVR) achieved the highest accuracy for NO$_{2}$ and SO$_{2}$ (R$^2$ = 0.71 for both), successfully capturing fine-scale spatial variability and multi-year temporal trends, including reductions associated with COVID-19 lockdowns. Predictions for PM$_{2.5}$ and CO showed moderate accuracy, whereas O$_3$ remained difficult to model.

\paragraph{Agriculture}

\cite{ma2025} assessed AlphaEarth Embeddings across three agricultural tasks in the United States: crop yield prediction, tillage mapping, and cover crop mapping. Using datasets from both public and private sources, they compared embedding-based approaches with conventional RS models. The results indicate that Foundation-based models achieve competitive performance for yield prediction and county-level tillage mapping when trained on local data, but exhibit limitations in spatial transferability, interpretability, and temporal sensitivity.
Focusing specifically on yield estimation, \cite{Fang2025} investigated whether
AlphaEarth Embeddings can serve as off-the-shelf features for county-scale yield estimation in the U.S. Corn Belt.
Using a single embedding per county–year and a lightweight radial basis function SVR model, they evaluated leave-one-year-out generalization. Despite the absence of time-resolved inputs, the embeddings achieved strong cross-year predictive performance, with mean test accuracies of R$^2 = 0.825$ ($RMSE = 14.8$ bu/ac) for corn, peaking at R$^2 = 0.861$ in the best-performing year, and R$^2 = 0.814$ for soybean. 
\cite{Murakami} examined the within- and cross-regional performance of crop classification using AlphaEarth across three cool climate agricultural regions: Northern Europe (Germany), Northern Japan (Hokkaido), and Northern US (Michigan). Within-region classification achieved high overall accuracy ($91.4-99.1\%$) irrespective of regions and classifiers (SVM, RF, and partial least squares). However, cross-regional transferability showed limited success. While models trained using the Japanese dataset exhibited high accuracy ($88.0-94.9\%$) in the US dataset, the reverse transfer and other combinations of transfer did not result in accurate classification. 
 
\paragraph{Biomass estimation}

\cite{Lucero2026} evaluated the potential of AlphaEarth Embeddings in estimating aboveground biomass in the Las Piedras River sub-basin (tropical Andean forests), covering an area of approximately 13 km$^2$. Specifically, they compared Embedding with conventional Sentinel-2 workflows based on spectral indices (e.g., NDVI, EVI, and SAVI). Using both RF and Artificial Neural Network (ANN) models, they found that spectral indices outperformed AlphaEarth Embeddings in this context. ANN models consistently surpassed RF, achieving a maximum classification accuracy of $79.0\%$. 

\paragraph{Hydrological Modeling}
\cite{Pengfei} showed that satellite-informed environmental representations can strengthen hydrological forecasting and support the development of models that adapt more easily to different landscapes.

\paragraph{Groundwater fluoride contamination} \cite{Wei2025} applied AlphaEarth Embeddings to predict groundwater fluoride contamination in the Datong Basin using RF, Support Vector Machine (SVM), and ANN models. Based on 391 groundwater samples, key predictors were selected using ANOVA F-values. The performance of the model was evaluated using ROC–AUC. The SVM model achieved the best performance (AUC = 0.82), outperforming RF (0.80) and ANN (0.77). Incorporating satellite embeddings improved all models, reducing prediction errors by 13.8–23.3\%. 

\paragraph{Canopy height regression}
\cite{feng2025tessera} introduced TESSERA embeddings and evaluated their performance in six downstream tasks and benchmarks, including classification, segmentation, and regression. Specifically, they focused on tree species classification, parcel segmentation, crop type classification, crop semantic segmentation, biomass regression, and canopy height regression. For each downstream task, they compared TESSERA with AlphaEarth, Presto, and traditional baselines. For canopy height regression, TESSERA, AlphaEarth, and Presto achieved Root Mean Square Error (RMSE) values of 12.2 m, 16.1 m, and 17.9 m, respectively, over Danum Valley in Borneo, covering approximately 
30 km$^2$. To our knowledge, this study represents the only existing evaluation of embedding-based approaches for height-related tasks. However, the task focuses on relative canopy height estimation for trees, rather than general surface height mapping, and on a very limited area. 

Table \ref{tab:lit_review} summarizes previous studies investigating AlphaEarth Embeddings. 

\begin{table}[!h]
\centering
\caption{Details of the area and resolution of the study for each downstream task investigated using AlphaEarth Embeddings.
\textsuperscript{*} The analysis is done in vectorial form and outside the native raster domain.}
\resizebox*{1\textwidth}{!}{
\begin{tabular}{llll}
\hline
Author & Task & Area & Resolution \\
\hline
\cite{Khan} & Land Use classification & 19.3 km$^2$ & 10 m \\
\cite{Alvarez2025} & Urban Air Quality prediction & 60 points\textsuperscript{*} & - \\
\cite{ma2025} & Agricultural tasks &  47709 points\textsuperscript{*}  & - \\
\cite{Fang2025} & Yield estimation & 992 points\textsuperscript{*}  & - \\
\cite{Murakami} & Crop classification & 270 points\textsuperscript{*} & - \\
\cite{Lucero2026} & Biomass estimation & 13 km$^2$ & 10 m\\
\cite{Pengfei} & Hydrological modeling & 671 points\textsuperscript{*} & - \\
\cite{Wei2025} & Groundwater fluoride contamination & 391 points\textsuperscript{*} & - \\
\cite{feng2025tessera} & Canopy height regression & 30 km$^2$ & 10 m\\
\hline
\textbf{Our study} & \textbf{Height Inference} & \textbf{7865 km$^2$} & \textbf{10 m} \\
\hline
\end{tabular}
}
\label{tab:lit_review}
\end{table}

In most cases (6 out of 9), AlphaEarth Embeddings were sampled at point locations and thus handled in vector form rather than processed in their native raster domain, primarily to reduce storage and computational demands. Even when raster-based approaches were adopted at the native 10 m resolution, analyses were typically limited to relatively small areas ($\leq$ 30 km$^2$), limiting spatial scalability. 
In contrast, our study leverages the uint8 (8-bit) version of the AlphaEarth Embeddings \citep{SC} and applies a dedicated normalization procedure (Section \ref{sec:preprocessing}), substantially reducing data volume and computational load. 
This strategy enabled direct raster-domain processing at 10 m resolution over a considerably larger extent ($7865$ km$^2$, Section \ref{sec:study_area}) than previous studies, thus overcoming their spatial limitations and demonstrating the feasibility of deploying AlphaEarth Embeddings at a regional scale.

\section{Study area and data}\label{sec:study_area}

\begin{figure}[h!]
    \centering
    \includegraphics[trim={0 0 0 3cm},clip, width=1\linewidth]{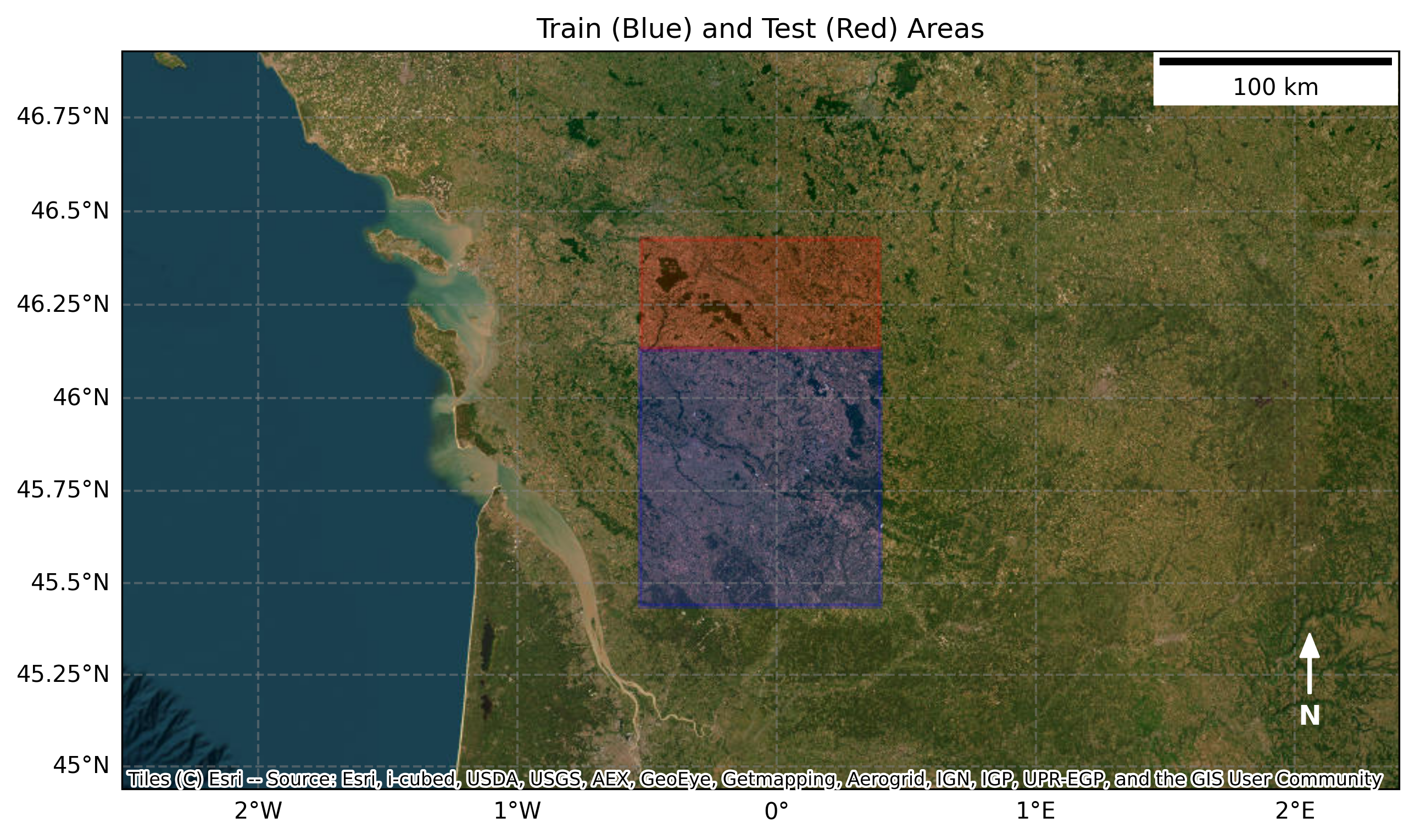}
    \caption{Study area: training and validation region in blue ($\approx5550$ km$^2$) and testing region in red ($\approx2315$ km$^2$).}
    \label{Fig1}
\end{figure}

The proposed methodology was evaluated in the Nouvelle-Aquitaine region (France), covering approximately $8000$ km$^2$, where a high-quality DSM was available.
To ensure spatial independence between training and testing samples, approximately 70\% of the area was used for training and validation, while the remaining contiguous 30\% was reserved for testing (Figure~\ref{Fig1}).

As input data, we used AlphaEarth Embeddings at 10 m spatial resolution \citep{brown2025AlphaEarthfoundationsembeddingfield, GoogleSatEmbeddingV1}, downloaded from Source Cooperative \citep{SC}, which provides the annual dataset (2018–2024) in Cloud-Optimized GeoTIFF (COG) format.
In this distribution, the embeddings are stored as signed 8-bit integers (\texttt{uint}, between -128 and 127) and therefore require normalization before use.
Encoding the AlphaEarth Embeddings with reduced numerical precision substantially decreases storage requirements compared with the GEE version, where embeddings are stored as double precision. As a result, download sizes are significantly smaller, facilitating also local processing.
Moreover, this setup enables the use of models and architectures that are not available within GEE. After normalization, users can select the desired floating-point representation (e.g., converting to float32 instead of float64), further optimizing memory usage while avoiding unnecessary computational overhead.

In our workflow, the AlphaEarth Embeddings served as input predictor variables for DL regression models aimed at estimating morphological information derived from a high-quality 5 $\times$ 5 m DSM provided by the French National Geographic Institute (IGN) \citep{IGN2021RGEALTI, IGN2021RGEALTI2}. The DSM, downloaded from GEE \citep{IGN2021RGEALTI2}, served as the target variable.
In particular, the DSM was used as reference data to train and evaluate the chosen DL regression models, enabling an assessment of how effectively the geospatial and multimodal information encoded in AlphaEarth Embeddings can support the prediction of terrain morphology.

For AlphaEarth Embeddings, we selected data from the year 2020, as it is the first year with a complete GEDI relative height record. The inclusion of GEDI data in the AlphaEarth Embeddings for that year is expected to improve the characterization of canopy height and facilitate the separation of vegetation structure from terrain height.
On the other hand, the IGN RGE ALTI 5 m DSM does not correspond to a single acquisition year. It is a seamless national product compiled from airborne LiDAR and photogrammetric surveys acquired over multiple years. In Nouvelle-Aquitaine, the underlying elevation data were collected at different times depending on the department, primarily between the late 2010s and early 2020s, and subsequently integrated into a unified DSM. As a result, the DSM provides a temporally aggregated yet spatially consistent reference of ground elevation across the study area \citep{IGN2021RGEALTI2}.

\section{Methodology}
The primary objective of this study is to investigate how geospatial and multimodal features encoded in AlphaEarth Embeddings can guide DL-based regression models to predict morphological information derived from high-quality regional DSMs.
To this end, we trained two DL regression models to map AlphaEarth Embeddings to the Nouvelle-Aquitaine DSM (Figure \ref{fig:training_pipeline}) and then applied the trained models to a previously unseen area (Figure \ref{fig:inference_pipeline}) (Section \ref{sec:Unet}). This approach allowed us to evaluate the versatility and transferability of embedding-driven DL regression models for height surface mapping, including a comparison with a linear baseline (Section \ref{sec:linear_baseline}).
The following sections describe the pre-processing step and the DL regression models in detail.

\subsection{Data pre-processing}\label{sec:preprocessing}

During preprocessing, pixels in the COG version of the AlphaEarth Embeddings with a no-data value of -128 were identified and reassigned to zero to prevent invalid values from affecting the learning process. In total, 1,043 out of 54,634,398 pixels were invalid in the training set (approximately 0.0019\%), while no invalid pixels were detected in the testing set.
The embeddings were subsequently normalized by adding 128 and dividing by 256, thereby rescaling the values to the [0, 1] range while preserving the relative structure of the original data

The IGN DSM was resampled to $10 \times 10$ m resolution using nearest-neighbor interpolation to match the AlphaEarth Embedding resolution. 
Finally, we stacked AlphaEarth Embedding with the DSM to construct input–target pairs, allowing the model to learn the relationship between embedding-derived features and corresponding surface height values.

\subsection{DL-based regression}\label{sec:Unet}

In our workflow, the 64 AlphaEarth Embedding bands served as input predictor variables for the DL-based regression models, while the high-resolution regional DSM provided the target variable to be predicted (Figure \ref{fig:training_inference_pipeline}, also the complete algorithm is detailed in \ref{appendix1}). 

\begin{figure}[ht!]

    \begin{subfigure}[b]{\linewidth}
    \centering    
    \includegraphics[width=0.75\linewidth]{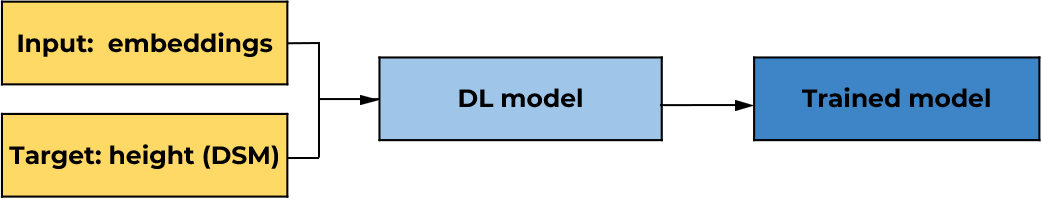}
    \caption{Training/validation pipeline.}
    \label{fig:training_pipeline}
    \end{subfigure}

    \begin{subfigure}[b]{\linewidth}
    \centering
    \includegraphics[width=1\linewidth]{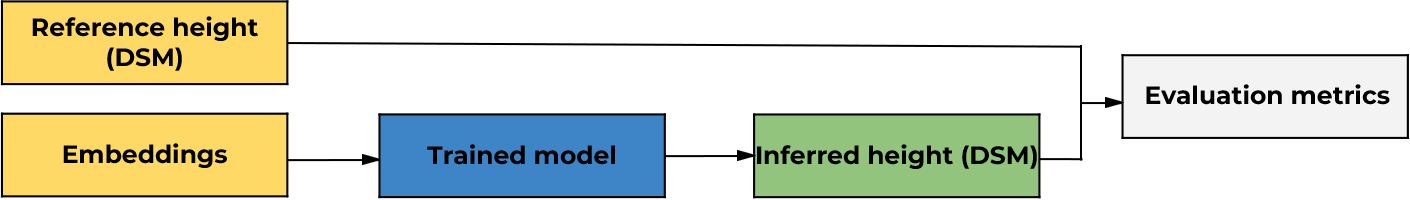}
    \caption{Testing pipeline (inference).}
    \label{fig:inference_pipeline}
    \end{subfigure}  
    \caption{Simplified overview of the proposed approach for training/validation (a) and testing (b).}
    \label{fig:training_inference_pipeline}
\end{figure}

We used U-Net \citep{Ronneberger} and U-Net++ \citep{Zhou2018} architectures, originally designed for image segmentation, to perform our regression task of inferring continuous height values at a spatial resolution of 10 m. Rather than adopting computationally intensive architectures (e.g., transformer-based models), we used lightweight U-Net convolutional encoder–decoder architectures as task-specific prediction heads \citep{feng2025tessera}.
Indeed, since AlphaEarth Embeddings are compact and semantically rich geospatial representations derived from massive pretraining, the regression network only needs to map embedding features to surface height values, without the need to perform complex feature extraction from raw data.
This choice thus leverages the rich geospatial information already encoded in the AlphaEarth Embeddings, enabling a clearer assessment of their height predictive value while allowing scalable regional applications thanks to substantially reduced computational requirements compared with training deep or attention-based models.

To ensure a fair and consistent comparison between U-Net and U-Net++, we trained both models using identical architectural settings, input data, and optimization parameters. 
For both models, we used a ResNet-18 encoder initialized with default ImageNet-pretrained weights.
Preliminary experiments with a deeper ResNet-50 encoder resulted in substantially longer training time without measurable performance gains. Therefore, we selected the shallower ResNet-18 backbone, as it provided comparable performance with greater computational efficiency.
We divided the input data into $512 \times 512$ patches, allowing the models to capture the spatial context of each area while also learning local patterns. 
We performed training using AdamW optimizer \citep{loshchilov2017decoupled} with an initial learning rate of $1 \times 10^{-3}$ and a weight decay of $1 \times 10^{-4}$ to improve generalization. 
We used the Mean Squared Error (MSE) loss as the objective function for regression, and we monitored the validation loss with a ReduceLROnPlateau scheduler, reducing the learning rate by a factor of 0.5 if no improvement was observed over 20 epochs. Early stopping with a patience of 50 epochs was applied to prevent overfitting.
We programmed the training for a maximum of 500 epochs, with model weights saved whenever a new best validation loss was reached.
During each epoch, batches from the training dataset were forward propagated, the MSE loss computed, and gradients backpropagated for parameter updates. 
Validation loss was computed at the end of each epoch without gradient updates to monitor generalization performance.
The main architectural choices and training parameters of the DL models are summarized in Table~\ref{tab:model_params}. 

\begin{table}[ht!]
\centering
\caption{Parameters and hyperparameters of the U-Net and U-Net++ models.}
\label{tab:model_params}
\begin{tabular}{ll}
\toprule
\textbf{Parameter} & \textbf{Value} \\
\midrule
Input data & AlphaEarth Embedding at native 10 m resolution\\
Target data & IGN RGE Alti DSM at 10 m resampled resolution\\
Architecture & U-Net / U-Net++ \\
Encoder & ResNet18 \\
Input channels & 64 \\
Output channels & 1 \\
Patch size & $512 \times 512$ \\
Batch size & 16 \\
Training/validation split & 80/20\% \\
Optimizer & AdamW \\
Learning rate & $1 \times 10^{-3}$ \\
Weight decay & $1 \times 10^{-4}$ \\
Loss function & Mean Squared Error (MSE) \\
Learning rate scheduler & ReduceLROnPlateau \\
Maximum number of epochs & 500 \\
Data shuffling & True (training only) \\
\bottomrule
\end{tabular}
\end{table}

At each epoch of the training process, we monitored the training and validation losses to assess the convergence and fluctuations of the learning rate.
At the end of the training, we selected the U-Net and U-Net++ models that achieved the lowest validation loss (Figure \ref{fig:unetloss}, Figure \ref{fig:upploss}, and \ref{appendix1}). 
We then applied the selected models to the training set and to the independent testing set, allowing evaluation of their inference performance against the reference DSM. In particular, we calculated the coefficient of determination ($R^2$), the Pearson correlation coefficient, and the Spearman correlation coefficient to quantify the agreement between the inferred and reference heights. 
Additionally, we computed the differences between the inferred and reference heights, according to Equation \ref{eq:difference}:

\begin{equation}
    \Delta H = H\textsubscript{p} - H\textsubscript{g}
    \label{eq:difference}
\end{equation}

where $H\textsubscript{p}$ represents the height inferred  from the proposed models and $H\textsubscript{g}$ denotes the corresponding ground-truth height from the reference DSM.
Finally, we evaluated a set of standard metrics on the height difference ($\Delta H$): mean, median, Standard Deviation (SD), RMSE, Normalized Median Absolute Deviation (NMAD), and the $25^{th}$ and $75^{th}$ percentiles \citep{Jacobsen, Hamoudzadehgedi31122025}.

\subsection{Comparison with a linear model baseline} \label{sec:linear_baseline}

As a baseline, we used a Ridge regression model \citep{Ridge} to predict surface heights from the same input features used for the DL models.
Ridge regression minimizes the sum of squared errors between predicted and observed values, while adding a penalty proportional to the square of the model coefficients \citep{mcdonald2009ridge}. 
This regularization encourages smaller, more stable coefficients, improving generalization on unseen data.
We trained the linear model baseline on the training dataset with the same inputs as the DL models. Then, we applied it to both the training and test regions. We used the same set of evaluation metrics for the U-Net and U-Net++ models to allow for direct comparison of predictive performance.

\section{Results and discussion}
\subsection{Training convergence}
\begin{figure}[ht!]
    \centering
    \includegraphics[trim={0 0 0 3cm}, clip, width=1\linewidth]{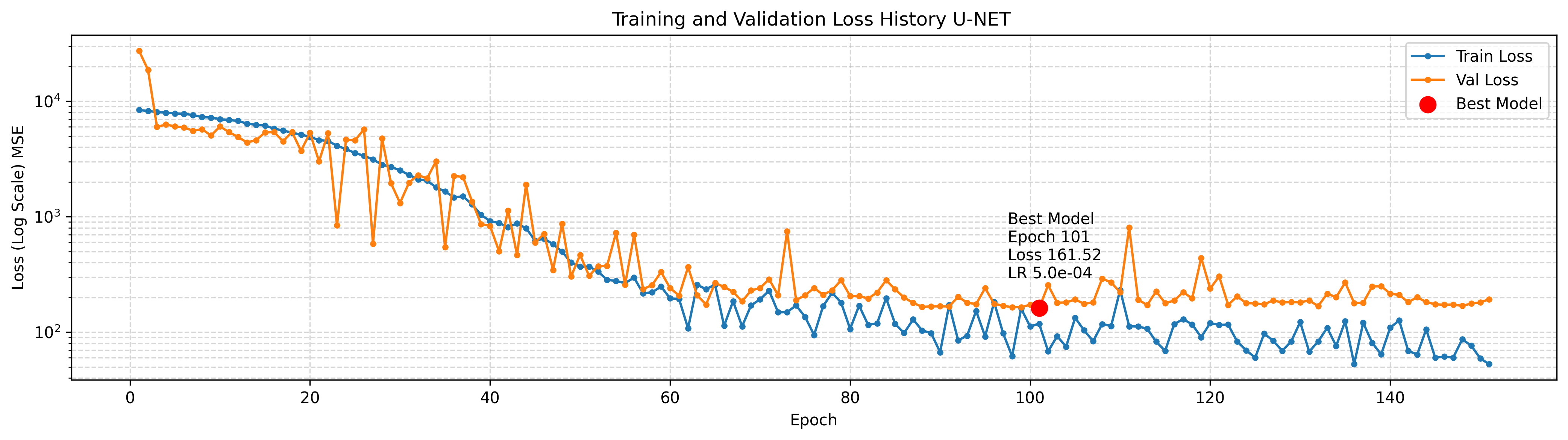}
    \caption{Training and validation losses and best model (U-Net) along with the corresponding loss, learning rate, and epoch number.}
    \label{fig:unetloss}
\end{figure}

The training of both U-Net and U-Net++ models showed steady loss convergence on the training and validation datasets. For the U-Net model, early stopping was triggered after 151 epochs. 
The training MSE decreased from an initial value of approximately 8000 m$^{2}$ to a range of approximately 50-120 m$^{2}$. 
The validation loss showed moderate fluctuations, reaching a minimum of 161.5 m$^{2}$ (corresponding to an RMSE of approximately 12.7 m) at epoch 101, which was selected as the best performing model (Figure \ref{fig:unetloss}).

\begin{figure}[h!]
    \centering
    \includegraphics[trim={0 0 0 3cm}, clip, width=1\linewidth]{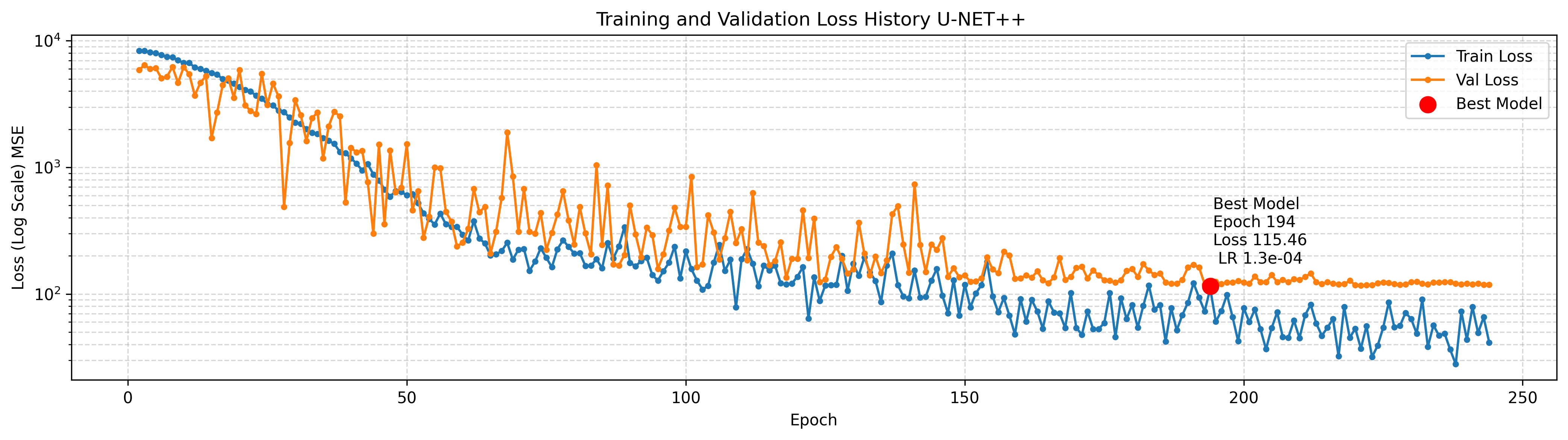}
    \caption{Training and validation losses and best model (U-Net++) along with the corresponding loss, learning rate, and epoch number.}
    \label{fig:upploss}
\end{figure}

For U-Net++, the training proceeded for 244 epochs before the early stopping was triggered. Compared to U-Net, U-Net++ consistently achieved lower training and validation losses, reaching a minimum validation loss of 115.5 m$^{2}$ (corresponding to an RMSE of approximately 10.7 m) at epoch 194 (Figure \ref{fig:upploss}).

In general, both models achieved stable convergence. However, U-Net++ exhibited more stable validation performance and lower final validation loss, indicating an improved generalization capability to unseen data.

\subsection{Model performance on the training set}

\begin{figure}[ht!]
    \centering
    \begin{subfigure}[b]{0.332\linewidth}
        \centering
        \includegraphics[width=\linewidth]{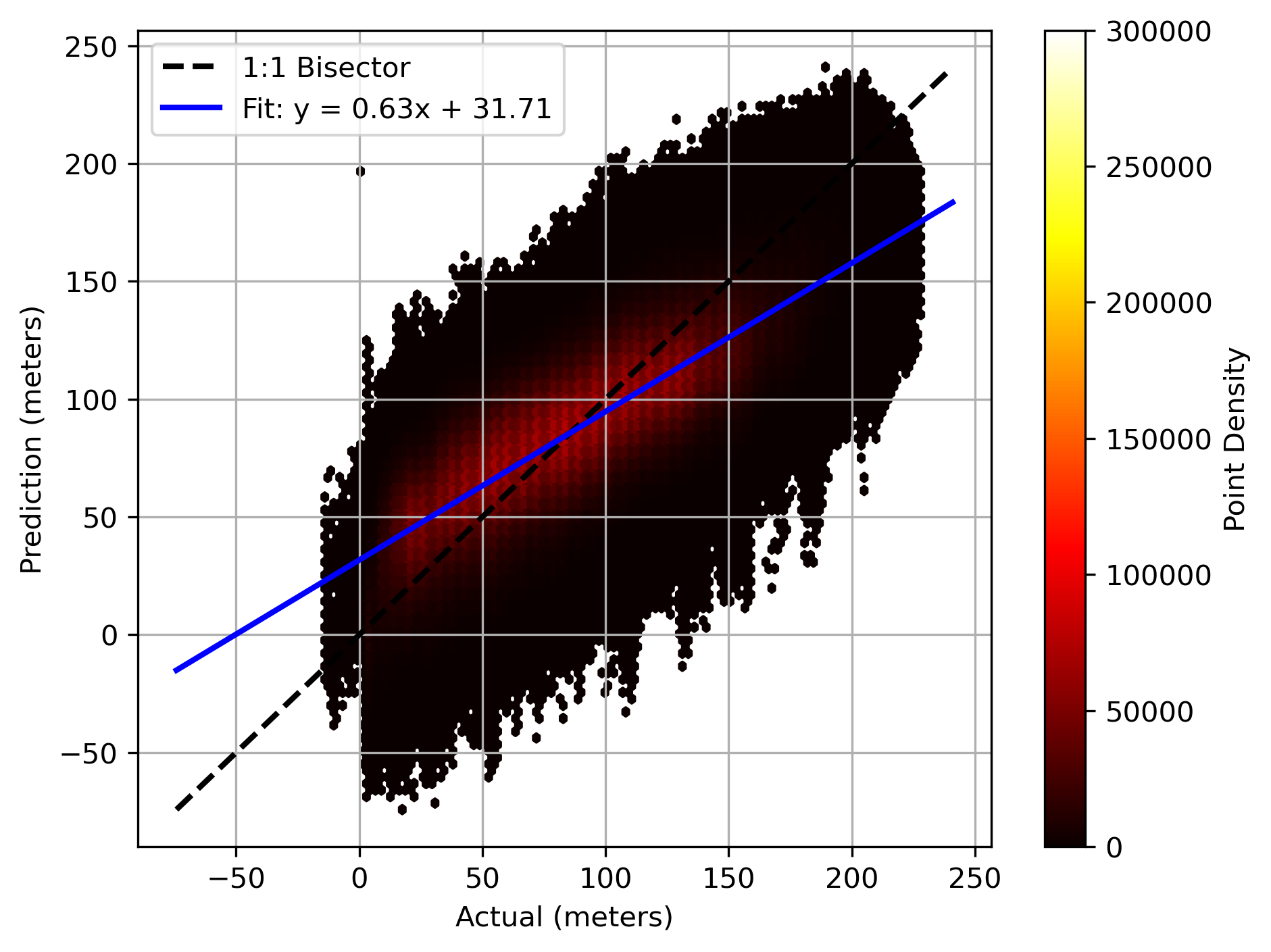}
        \caption{Ridge regression}
    \label{fig:ridge_training_scatter}
    \end{subfigure}
    \hfill
    \begin{subfigure}[b]{0.32\linewidth}
        \centering
        \includegraphics[width=\linewidth]{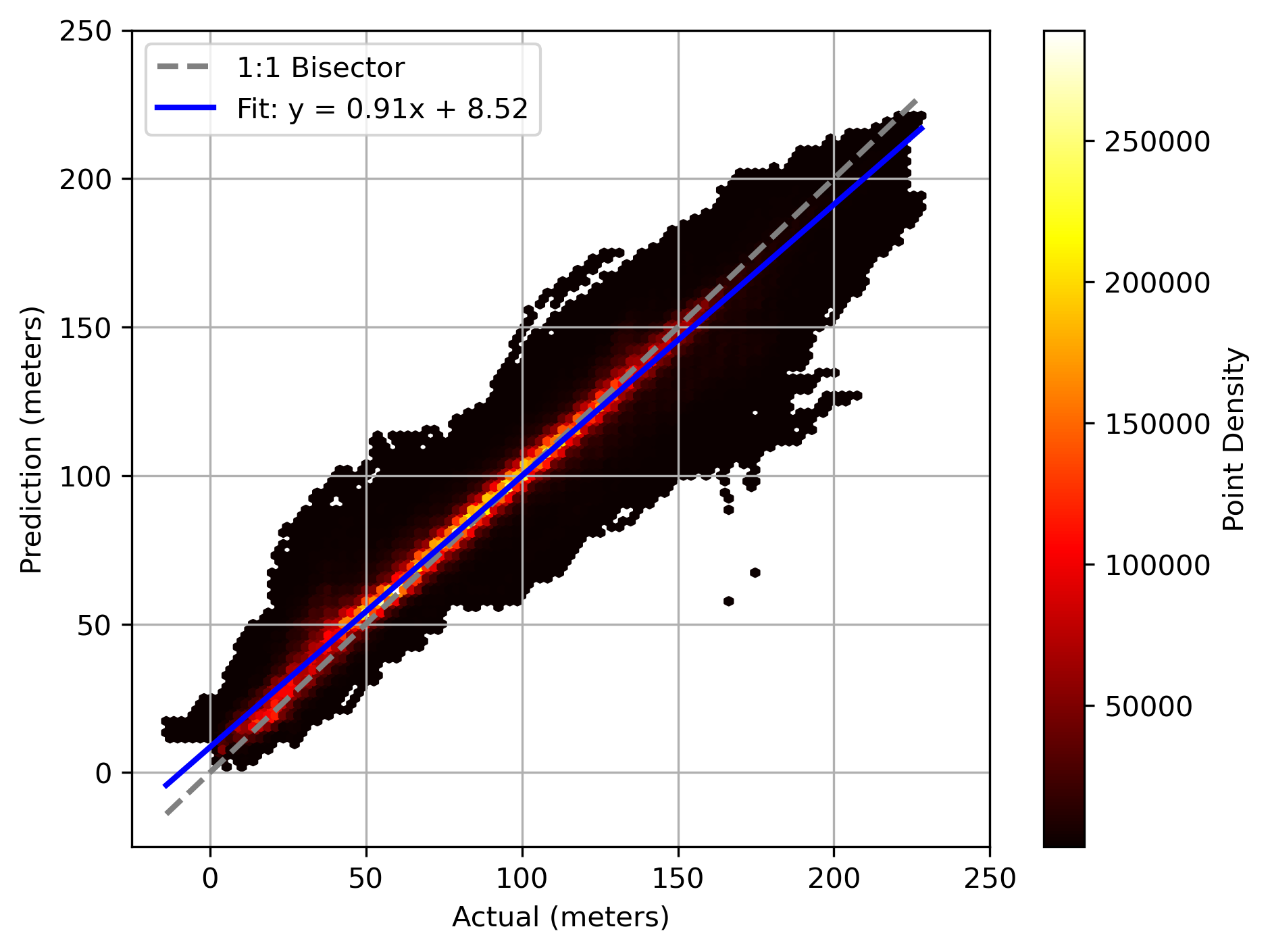}
        \caption{U-Net}
        \label{fig:unet_train_scatter}
    \end{subfigure}
    \hfill
    \begin{subfigure}[b]{0.32\linewidth}
        \centering
        \includegraphics[width=\linewidth]{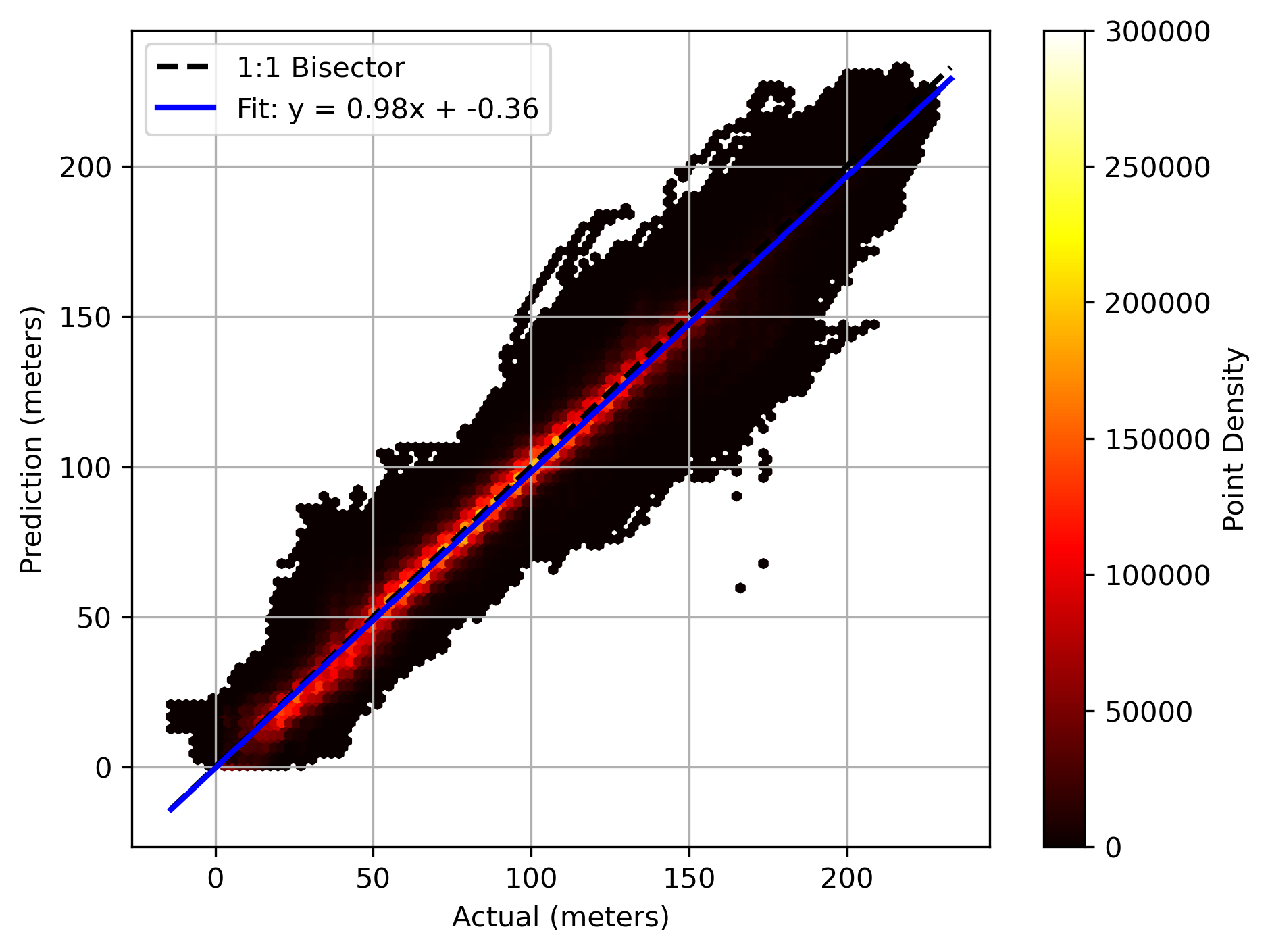}
        \caption{U-Net++}
        \label{fig:upp_histo_scatter}
    \end{subfigure}

    \caption{Correlation between inferred height and reference height for each model (training). The black diagonal line y = x represents the best possible fit; the blue line is the actual fit to the plot.}
    \label{fig:training_scatter}
\end{figure}

The U-Net and U-Net++ models that achieved the lowest validation loss during training were selected for subsequent quantitative and qualitative analyses. 
Figure \ref{fig:training_scatter} presents the pixel-wise correlation between the inferred and reference heights over the training dataset for each investigated model: Ridge regression (Figure \ref{fig:ridge_training_scatter}), U-Net (Figure \ref{fig:unet_train_scatter}), and U-Net++ (Figure \ref{fig:upp_histo_scatter}).

For Ridge regression, the fitted curve (blue line) deviates from the diagonal reference line (black), indicating a systematic tendency to overestimate heights below 100 m and underestimate heights above 100 m. 
In contrast, height predictions from U-Net and U-Net++ closely follow the diagonal line, reflecting stronger agreement with the reference heights.
Among the evaluated DL models, U-Net++ achieved the best performance, with the highest coefficient of determination (R² = 0.97), Pearson correlation (0.99), and Spearman correlation (0.99), as well as the lowest bias (–0.36 m). A summary of the quantitative training results for all models is provided in Table \ref{tab:train_test_results}.

\begin{figure}[h!]
    \centering

    \begin{subfigure}{0.49\textwidth}
        \centering
        \includegraphics[trim={0 0 0 0.66cm},clip, width=\textwidth]{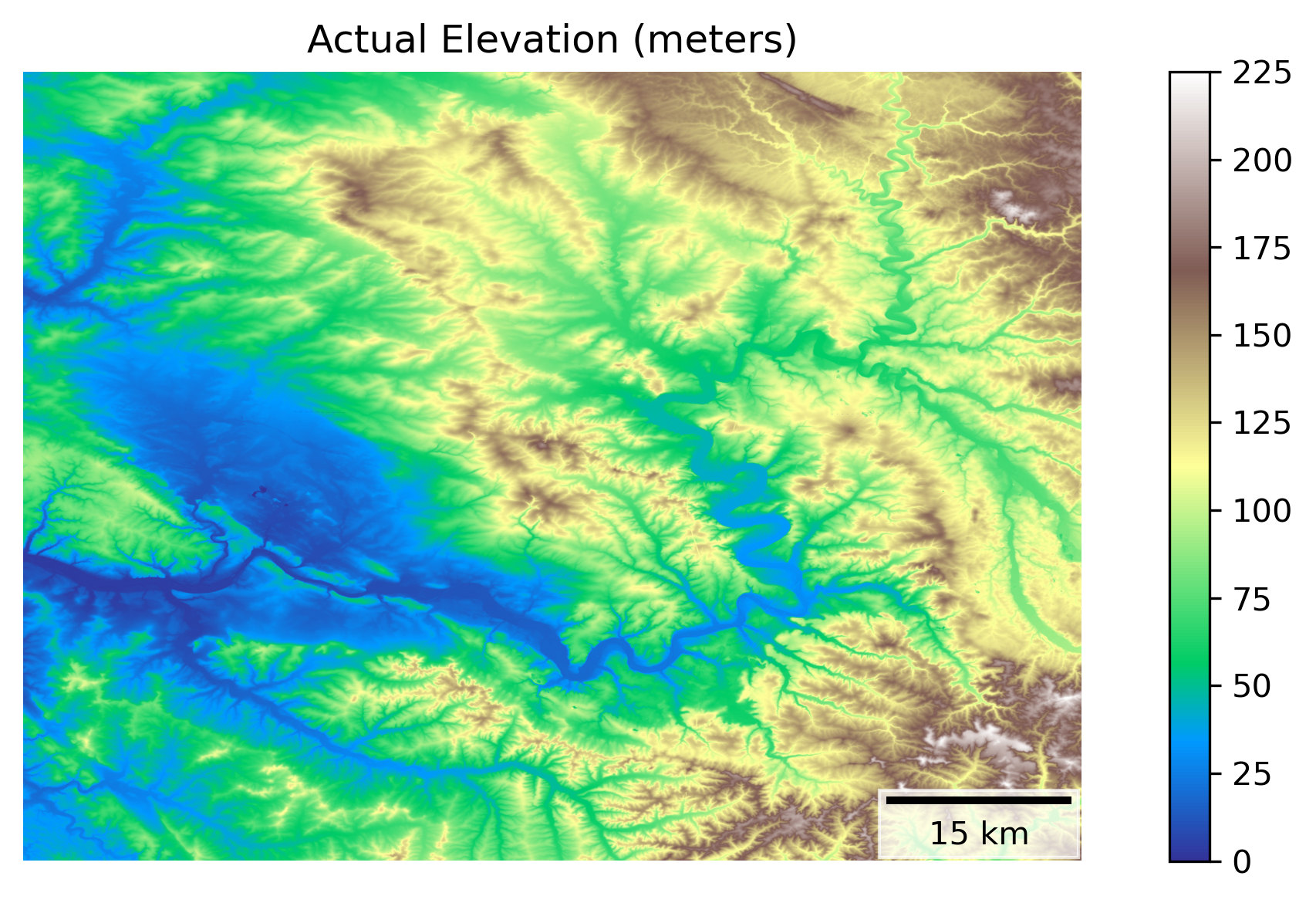}
        \caption{Reference height}
        \label{fig:reference_height}
    \end{subfigure}
    \hfill
    \begin{subfigure}{0.49\textwidth}
        \centering
        \includegraphics[trim={0 0 0 0.67cm},clip, width=\textwidth]{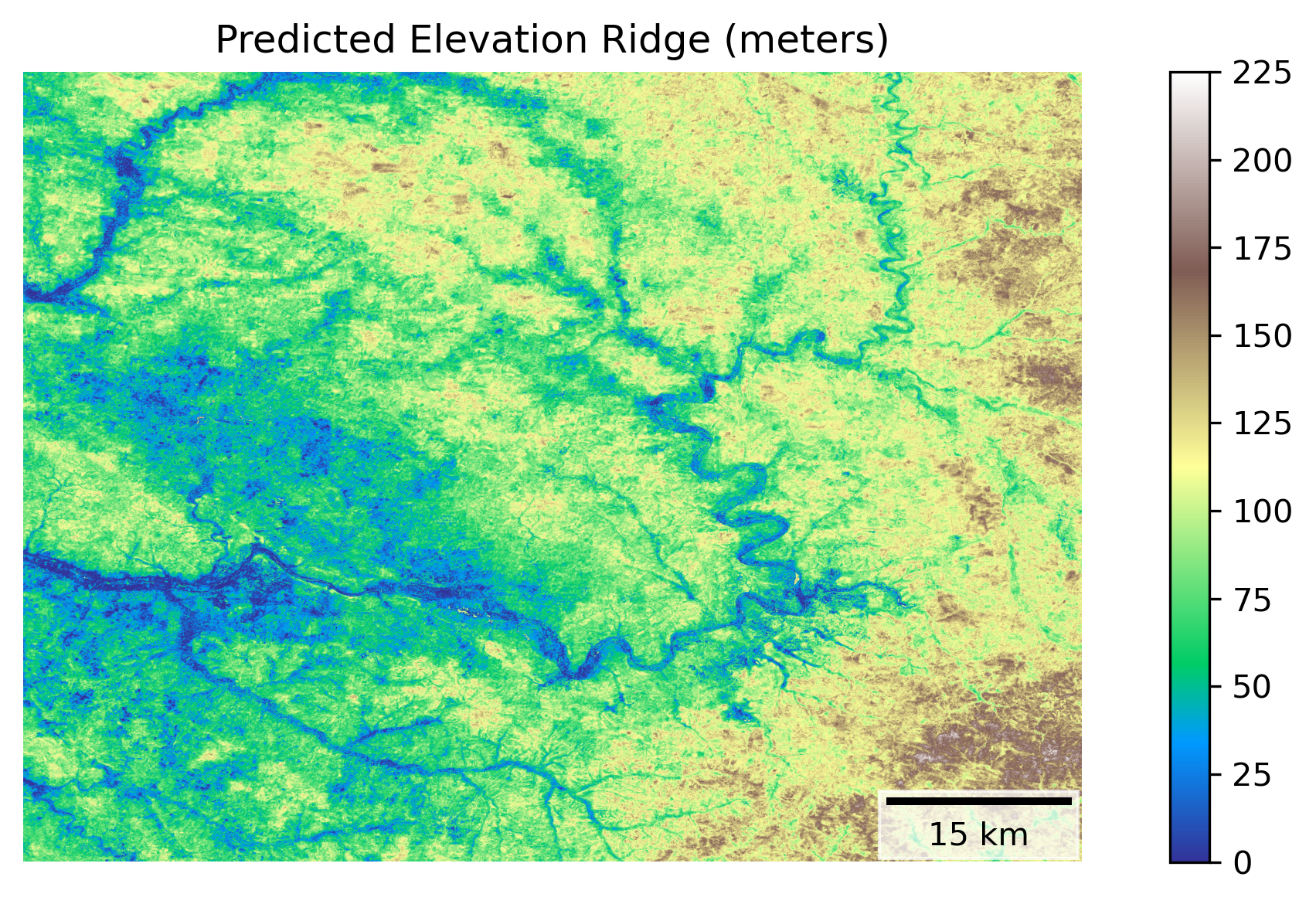}
        \caption{Height inferred from Ridge regression}
        \label{fig:ridge_height}
    \end{subfigure}

    \vspace{0.3cm}

    \begin{subfigure}{0.49\textwidth}
        \centering
        \includegraphics[trim={0 0 0 0.66cm},clip, width=\textwidth]{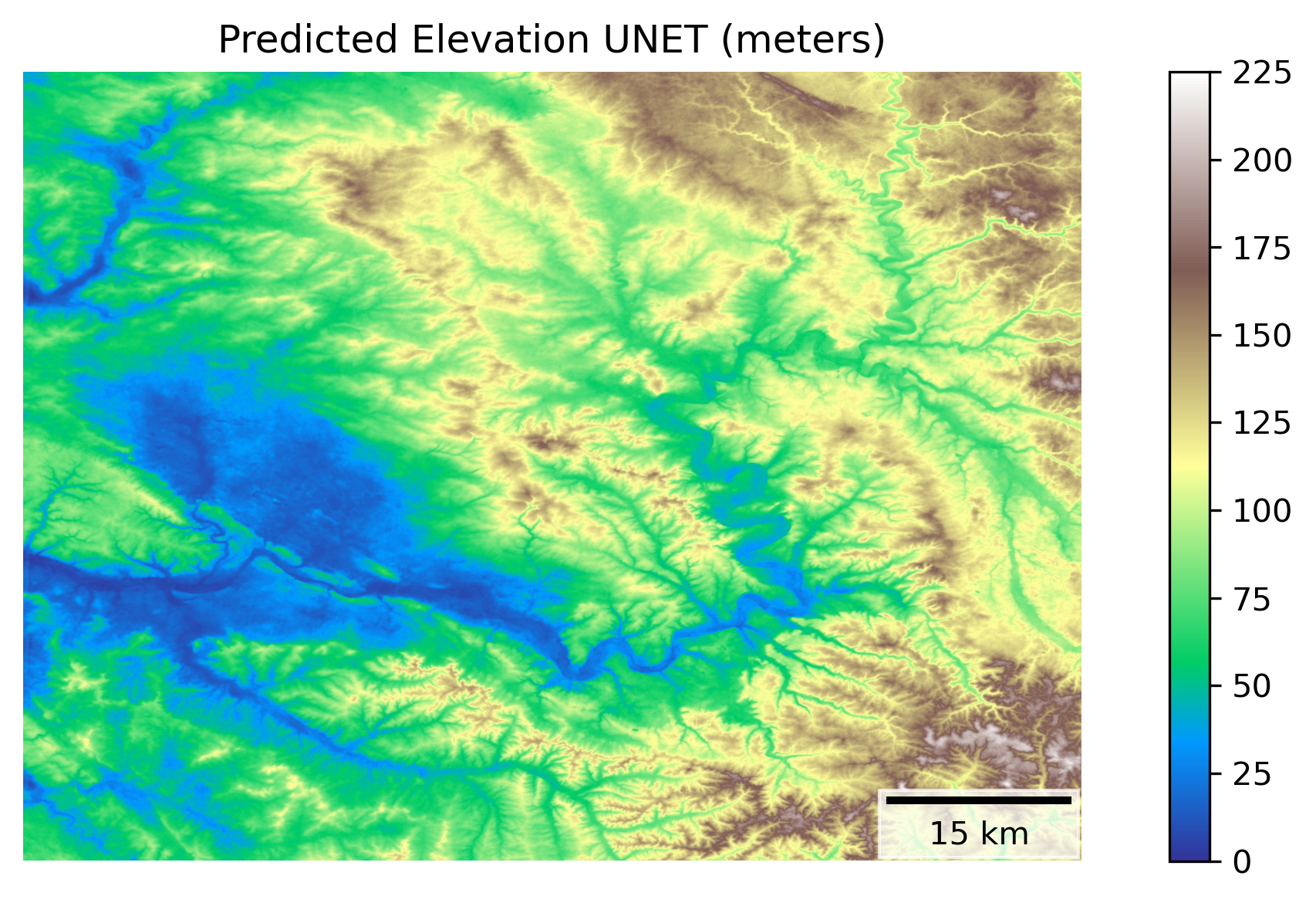}
        \caption{Height inferred from U-Net}
        \label{fig:unet_height}
    \end{subfigure}
    \hfill
    \begin{subfigure}{0.49\textwidth}
        \centering
        \includegraphics[trim={0 0 0 0.66cm},clip, width=\textwidth]{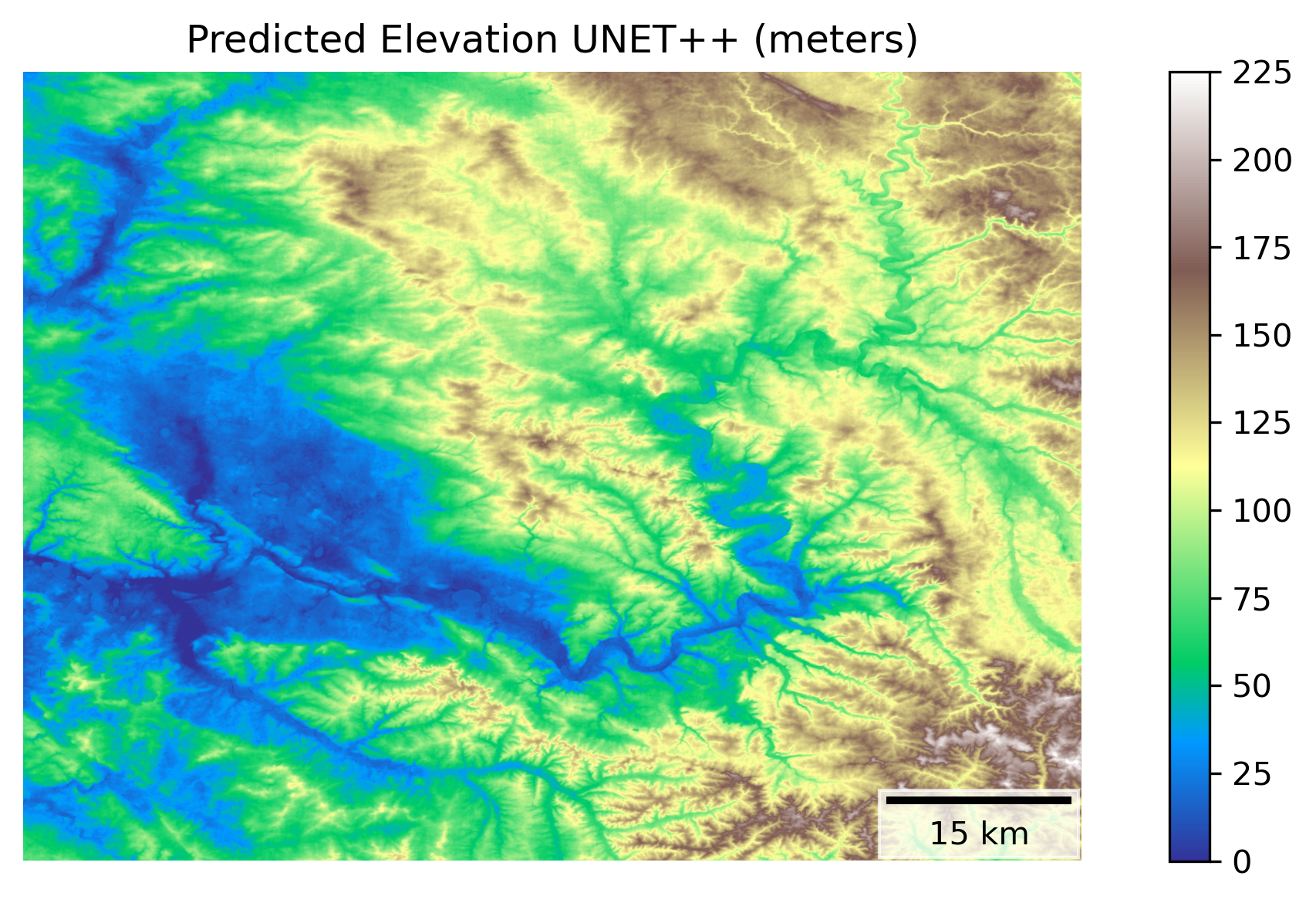}
        \caption{Height inferred from U-Net++}
        \label{fig:unet++_height}
    \end{subfigure}

    \caption{Qualitative results: Reference height and inferred height (meters) from the tested models (training).}
    \label{fig:training}
\end{figure}

Figure \ref{fig:training} shows the qualitative results in the training area, including the reference DSM (Figure \ref{fig:ridge_height}), Ridge regression (Figure \ref{fig:ridge_height}), U-Net (Figure \ref{fig:unet_height}) and U-Net++ (Figure \ref{fig:unet++_height}) inferred heights.
Consistent with the quantitative analysis, Ridge regression exhibited the weakest reconstruction performance, whereas U-Net and U-Net++ provided more accurate and spatially coherent height estimates.

Table \ref{tab:train_test_results} reports the evaluation metrics computed from the height differences ($\Delta H$) between the reference and the inferred values for the training area.

\subsection{Model performance on the testing set}

\begin{figure}[ht!]
    \centering
    \begin{subfigure}[b]{0.32\linewidth}
        \centering
        \includegraphics[trim={0 0 0 0.69cm},clip, width=\linewidth]{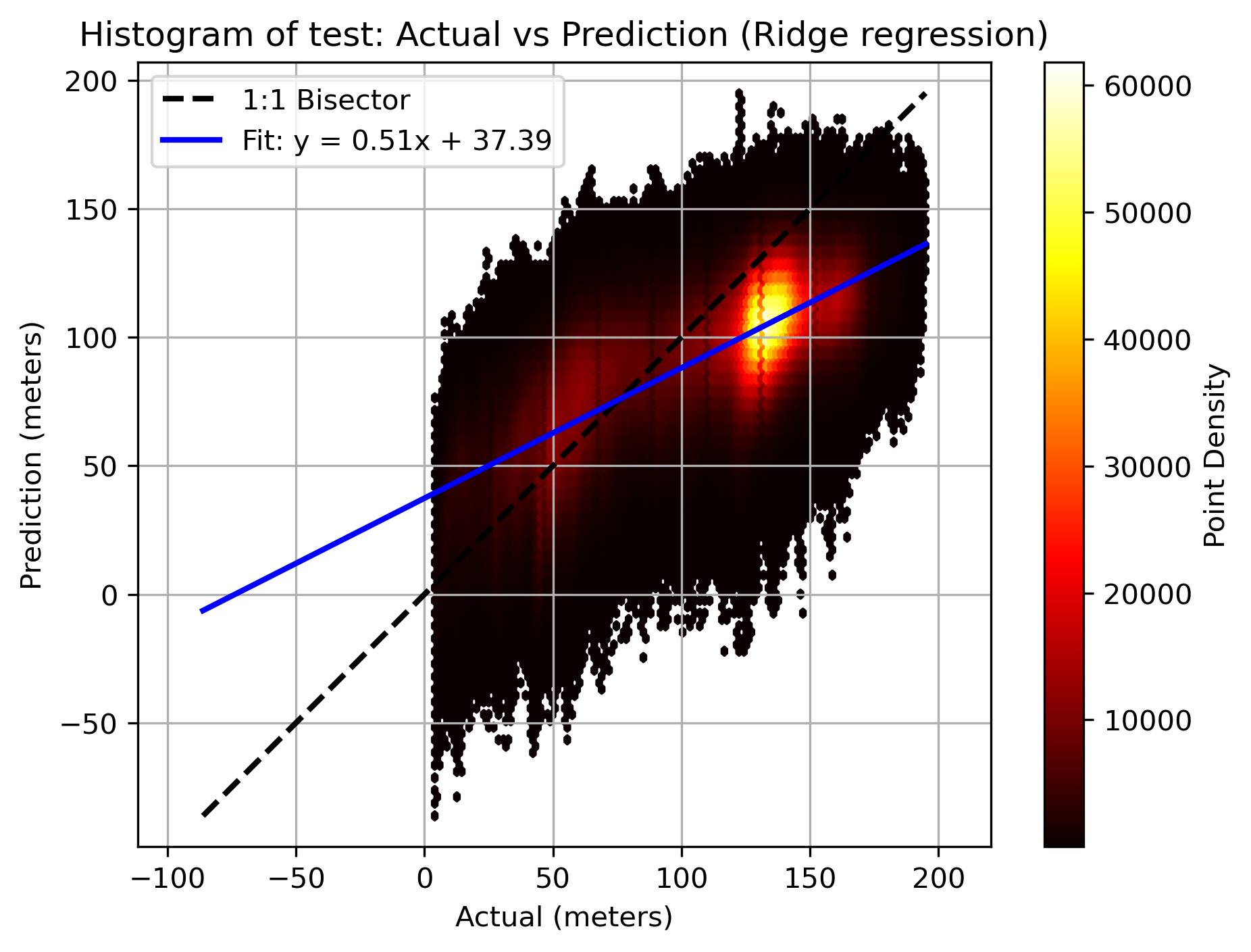}
        \caption{Ridge regression}
        \label{fig:ridge_scatter}
    \end{subfigure}
    \hfill
    \begin{subfigure}[b]{0.32\linewidth}
        \centering
        \includegraphics[trim={0 0 0 0.64cm},clip, width=\linewidth]{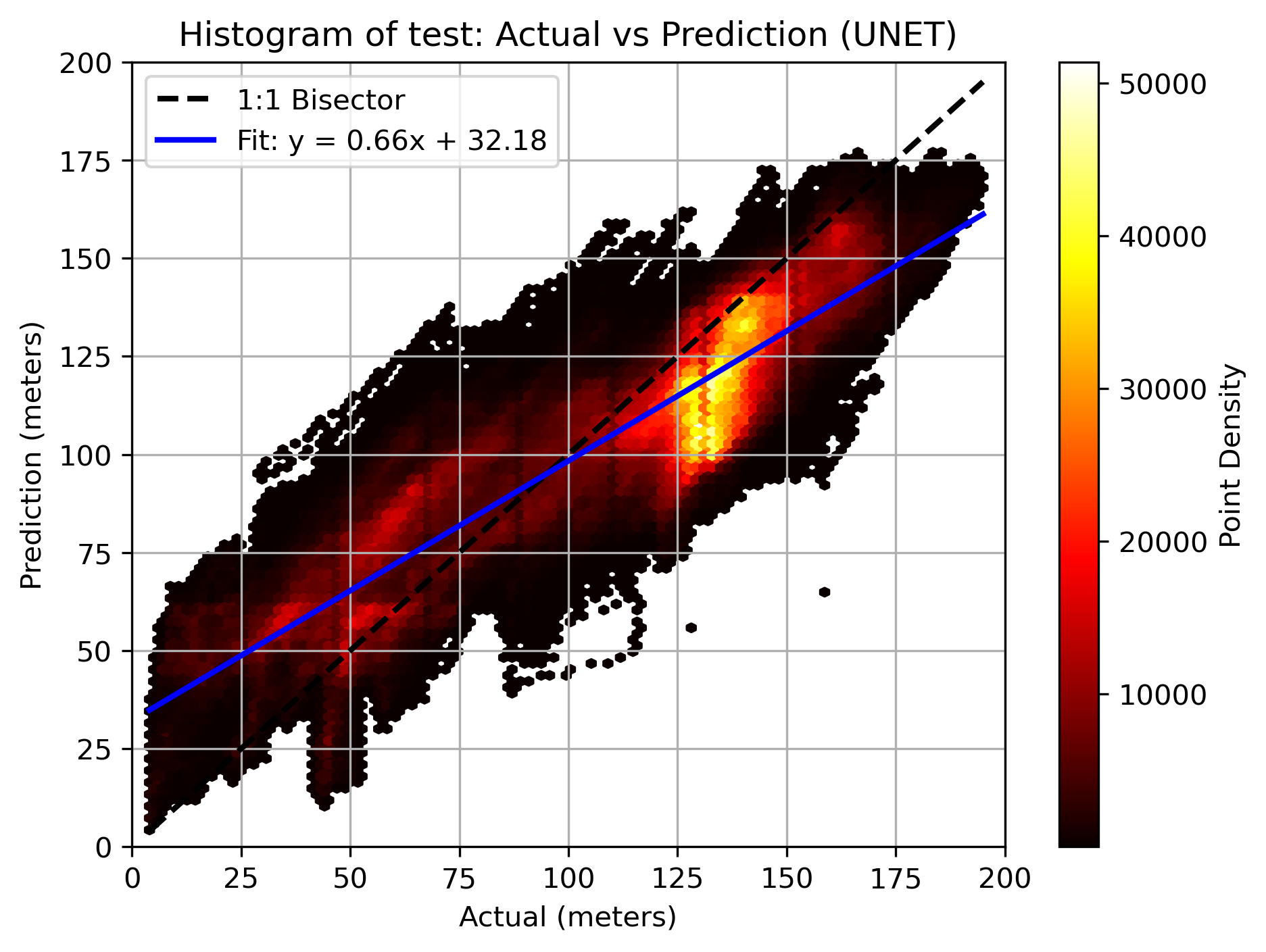}
        \caption{U-Net}
        \label{fig:unet_scatter}
    \end{subfigure}
\hfill
    \begin{subfigure}[b]{0.32\linewidth}
        \centering
        \includegraphics[trim={0 0 0 0.64cm},clip, width=\linewidth]{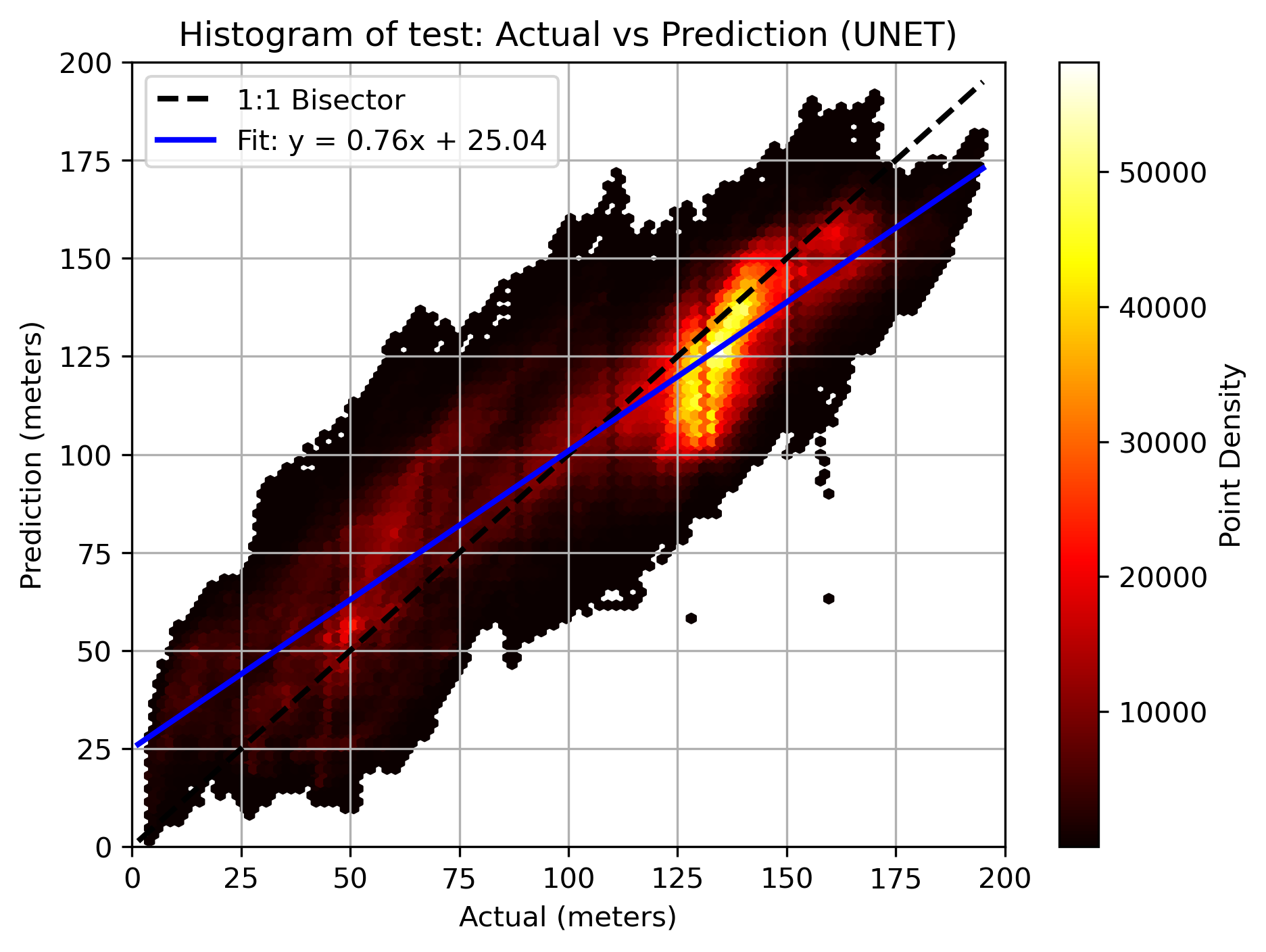}
        \caption{U-Net++}
        \label{fig:upp_scatter}
    \end{subfigure}

    \caption{Correlation between inferred height and reference height for each model (testing). The black diagonal plot y=x represents the best possible fit; the blue line is the actual fit to the plot.}
    \label{fig:testing_scatter}
\end{figure}

As a final step, we evaluated the model inference performance on the independent testing dataset. 
Figure \ref{fig:testing_scatter} shows the pixel-wise correlation on the testing set between inferred and reference heights for the Ridge regression baseline (Figure \ref{fig:ridge_scatter}), U-Net (Figure \ref{fig:unet_scatter}), and U-Net++ (Figure \ref{fig:upp_scatter}) models.
All models exhibited a systematic tendency to overestimate heights below 100 m and underestimate heights above 100 m. Ridge regression showed the weakest performance, while U-Net++ achieved the best results, with an R² of 0.84, Pearson correlation of 0.92, Spearman correlation of 0.91, and an average bias of approximately 25 m (Figure \ref{fig:testing_scatter}, Table \ref{tab:train_test_results}). 
These results indicate that AlphaEarth Embeddings show promising potential to guide DL-based height mapping workflows, enabling U-Net and U-Net++ models to reproduce surface heights in strong agreement with the reference data.

\begin{figure*}[ht!]
    \centering

    \begin{subfigure}{0.49\textwidth}
        \centering
        \includegraphics[width=\textwidth]{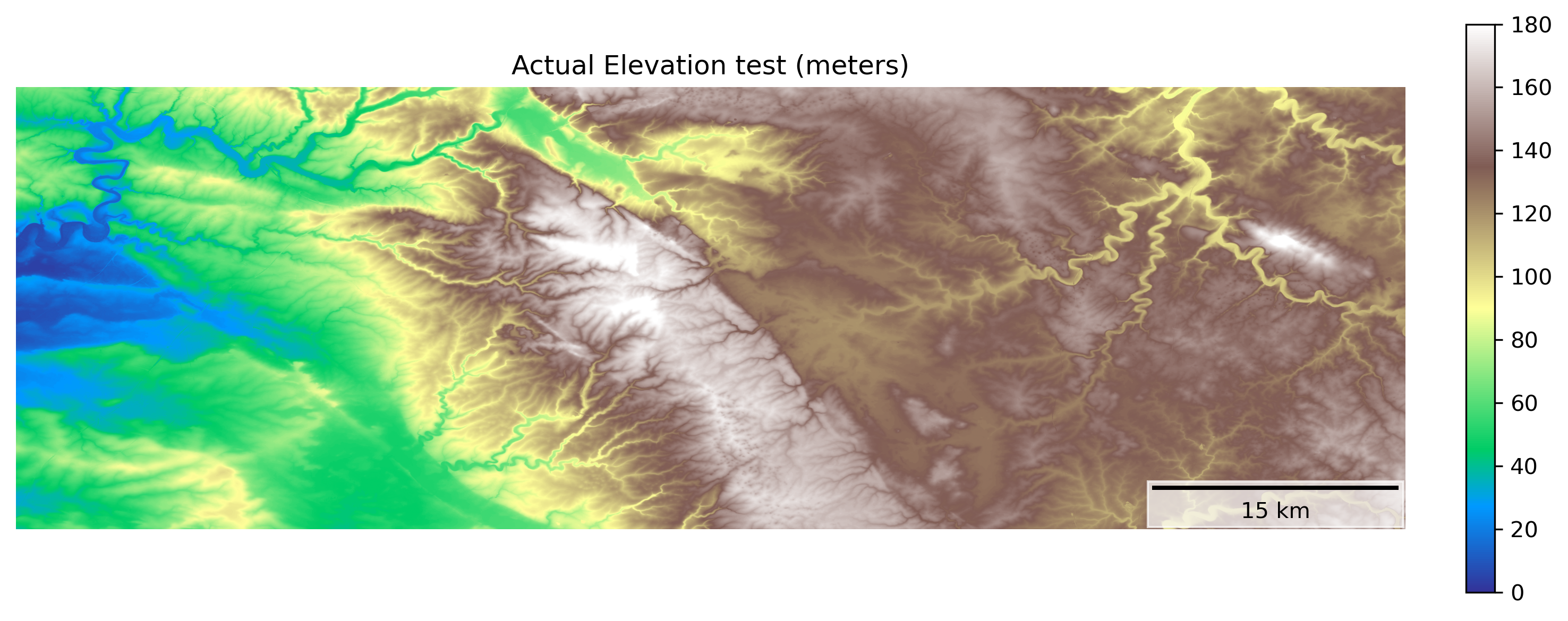}
        \caption{Reference height}
    \end{subfigure}
    \hfill
    \begin{subfigure}{0.49\textwidth}
        \centering
        \includegraphics[width=\textwidth]{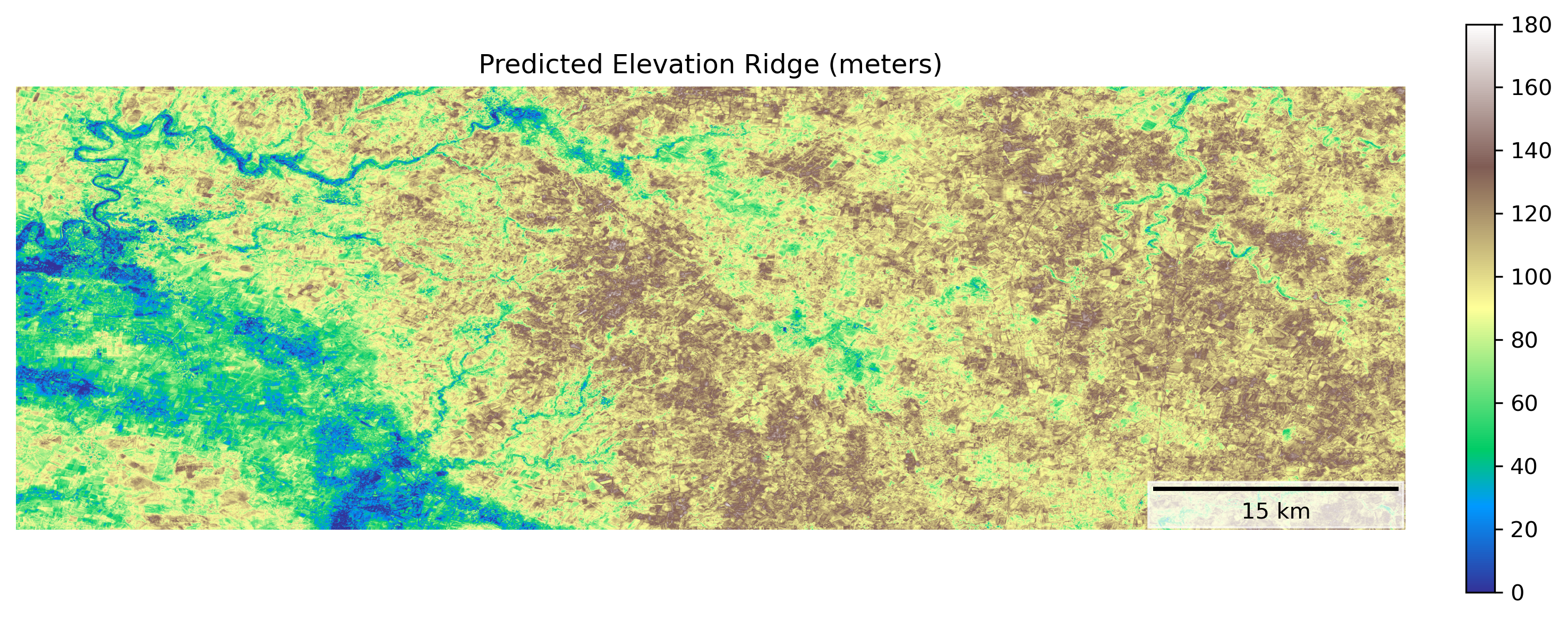}
        \caption{Height inferred from Ridge regression}
    \end{subfigure}

    \vspace{0.3cm}

    \begin{subfigure}{0.49\textwidth}
        \centering
        \includegraphics[width=\textwidth]{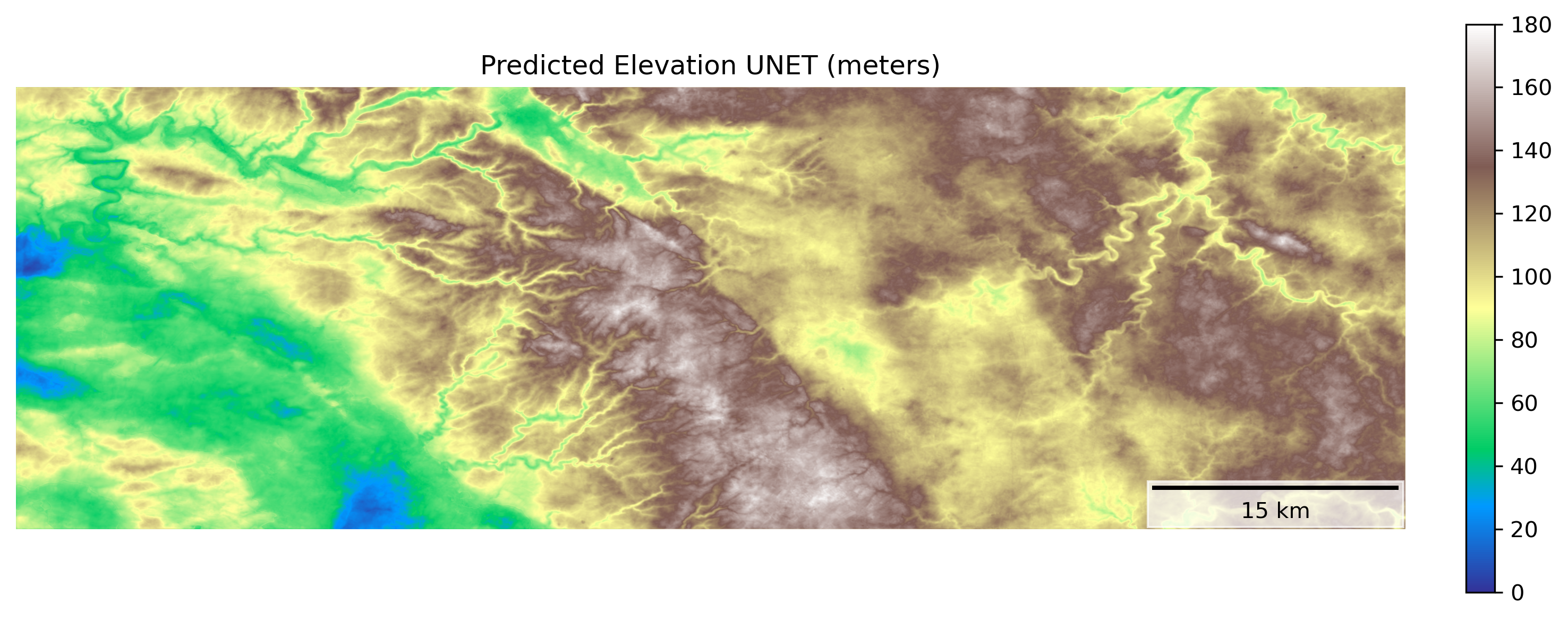}
        \caption{Height inferred from U-Net}
    \end{subfigure}
    \hfill
    \begin{subfigure}{0.49\textwidth}
        \centering
        \includegraphics[width=\textwidth]{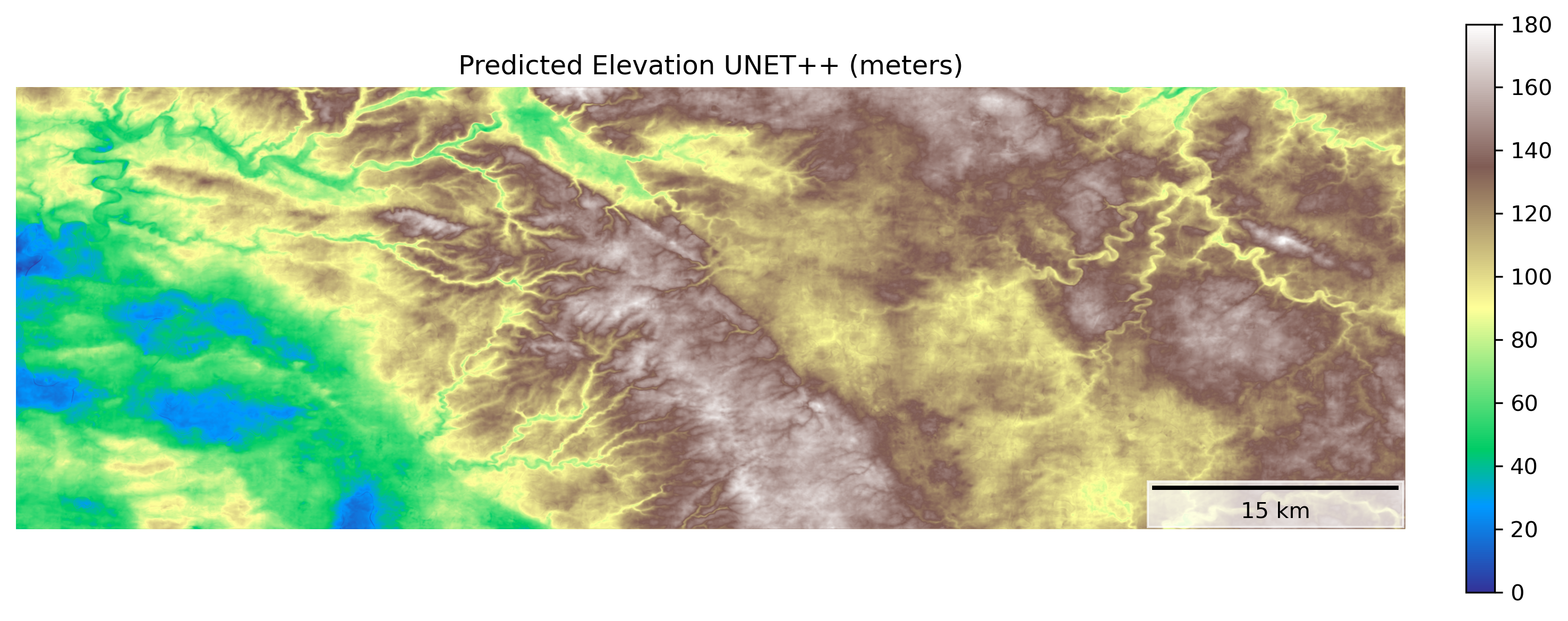}
        \caption{Height inferred from U-Net++}
    \end{subfigure}

    \caption{Qualitative results: Reference height and inferred height (meters) from the tested models (testing).}
    \label{fig:testing_qualitative}
\end{figure*}

Figure \ref{fig:testing_qualitative} presents qualitative results for the three models in the testing area, while Table \ref{tab:train_test_results} summarizes the corresponding evaluation metrics. Overall, the results confirm that U-Net++ achieved the best performance, with an RMSE of approximately 16 m on the testing set.

\begin{table}[ht!]
\centering
\caption{Model performance and difference statistics for U-Net, U-Net++, and Ridge Regression on training and test datasets.
All statistics are computed between the inferred heights and the reference DSM. }
\label{tab:train_test_results}
\begin{tabular}{lcccccc}
\toprule
\textbf{} 
& \multicolumn{2}{c}{\textbf{U-Net}} 
& \multicolumn{2}{c}{\textbf{U-Net++}} 
& \multicolumn{2}{c}{\textbf{Ridge Regression}} \\
\cmidrule(lr){2-3} \cmidrule(lr){4-5} \cmidrule(lr){6-7}
& \textbf{Train} & \textbf{Test} 
& \textbf{Train} & \textbf{Test} 
& \textbf{Train} & \textbf{Test} \\
\midrule
\textbf{Performance metrics} & & & & & & \\
\midrule
$R^{2}$ & 0.97 & 0.78 & 0.97 & 0.84 & 0.64 & 0.38 \\
Pearson correlation & 0.99 & 0.91 & 0.99 & 0.92 & 0.80 & 0.73 \\
Spearman correlation & 0.99 & 0.90 & 0.99 & 0.91 & 0.81 & 0.69 \\
\midrule
\textbf{Difference metrics} & & & & & & \\
\midrule
Mean difference (m) & 1.15 & -4.72 & -1.66 & -1.31 & 0.00 & -16.42 \\
Median difference (m) & 1.25 & -7.22 & -1.62 & -2.62 & 0.68 & -19.68 \\
SD (m) & 7.58 & 18.67 & 7.10 & 16.37 & 25.02 & 27.73 \\
RMSE (m) & 7.67 & 19.26 & 7.29 & 16.42 & 25.02 & 32.22 \\
NMAD (m) & 5.11 & 17.92 & 5.63 & 15.56 & 25.57 & 27.40 \\
25$^{th}$ percentile (m) & -2.17 & -18.36 & -5.39 & -12.97 & -16.98 & -36.15 \\
75$^{th}$ percentile (m) & 4.73 & 6.29 & 2.19 & 8.04 & 17.53 & 1.81 \\
\bottomrule
\end{tabular}

\end{table}

\subsection{Model performance and generalization}
The results reported in Table \ref{tab:train_test_results} show that DL architectures clearly outperform the linear Ridge regression baseline for surface height inference. Both U-Net and U-Net++ achieved very high training performance ($R^2 = 0.97$), indicating that AlphaEarth Embeddings encode informative topographic signals that can be effectively decoded by spatially aware models.

In the test set, performance decreased for all models, reflecting the shift in height frequency distribution between training and testing regions (Section \ref{sec:height_distribution}). 
Nevertheless, U-Net++ demonstrated better generalization ($R^2 = 0.84$) compared with the standard U-Net ($R^2 = 0.78$), suggesting that its nested skip connections contribute to better preservation of multi-scale spatial context and reduced overfitting. 
In contrast, Ridge regression exhibited a substantial performance gap between training and testing ($R^2 = 0.64$ and $R^2 = 0.38$, respectively), highlighting the limitations of purely linear decoding in capturing the complex, non-linear relationships between embedding features and surface height.

U-Net++ also achieved better difference statistics ($\Delta H$), with lower RMSE and NMAD values than both U-Net and Ridge regression in the training and testing datasets, indicating more stability under heterogeneous terrain conditions. In particular, the relatively close agreement of U-Net++ between mean and median differences (e.g., -1.66 m vs. -1.62 m on training data, and -1.31 m vs. -2.62 m on testing data) with respect to the other two models suggested a limited influence of outliers and a more balanced error distribution. By contrast, the larger discrepancies reflected in the residual values for Ridge regression, particularly in the testing set, indicated skewness in the error distribution. Specifically, the mean and median residuals are -16.42 and -19.68, respectively, both very far from zero, compared to the U-Net and U-Net++ models. The smaller dispersion metrics (SD, interquartile range, and NMAD) observed for U-Net++ further support its increased robustness, reinforcing the suitability of DL architectures for height mapping under heterogeneous terrain conditions.

Beyond aggregate error statistics, the models differ markedly in terms of the physical plausibility of their predictions. Analysis of the correlation scatter plots shows that neither U-Net nor U-Net++ produces negative height values in either the training or testing datasets. In contrast, the Ridge regression baseline predicts markedly negative heights, reaching approximately -80 m for training and -90 m for testing (Figures \ref{fig:ridge_training_scatter} and Figure \ref{fig:ridge_scatter}).
These extreme negative values are not supported by the reference DSM and therefore represent physically implausible height values for the study area. This behavior aligns with the large negative bias and broad error dispersion observed in the Ridge regression statistics, indicating unstable extrapolation under distribution shift (Section \ref{sec:height_distribution}).
By contrast, U-Net-based models generate predictions within realistic height ranges. This suggests that the combination of spatially structured decoding and embedding-derived geospatial features contributes to more physically consistent height estimates, even without explicitly imposed output constraints.

\subsection{Influence of height distribution in training and testing datasets}\label{sec:height_distribution}

\begin{figure}
    \centering
    \includegraphics[width=1\linewidth]{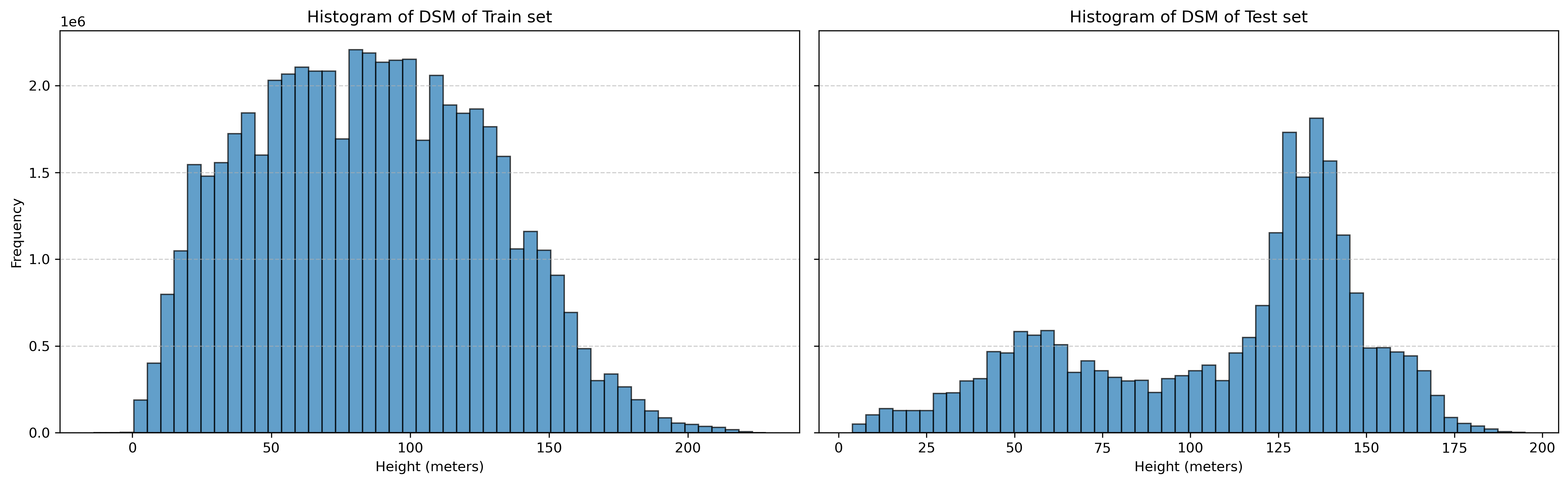}
    \caption{Histogram of the height distribution of DSM over the test and train sets.}
    \label{fig:dsm_histo}
    
\end{figure}

The height distributions of the training and testing datasets cover a similar overall range but differ in frequency. The training dataset is dominated by heights between 0 and 150 m, whereas the testing data contain a higher proportion of pixels in the 110–160 m range (Figure \ref{fig:dsm_histo}). This distribution shift likely contributes to the observed decrease in model performance on the testing set, as reflected by the higher error metrics.

Despite the mismatch in height distributions, U-Net++ achieves relatively strong agreement with reference heights, with a Pearson correlation of 0.92 and an RMSE of approximately 16 m on the testing set. 
These results demonstrate that the semantically rich AlphaEarth Embeddings can guide DL models to capture key height patterns even under shifts in height frequency. In this context, the architectural design of U-Net++ further contributes to robustness against non-uniform distributions. 
However, while the height predictions guided by AlphaEarth Embeddings exhibit strong correlation and suggest promising potential for DL-based height mapping workflows, the high RMSE on the testing set underscores the difficulty of achieving accurate generalization across regions characterized by different height distributions.
Nevertheless, our results are consistent with previous embedding-based studies. \cite{feng2025tessera} reported an RMSE of 16.1 m for canopy height regression using AlphaEarth Embeddings over a relatively limited study area (approximately 30 km$^2$), where maximum canopy heights did not exceed 80 m. In comparison, our downstream height inference over a substantially larger test region achieved a similar RMSE of 16.4 m, despite heights reaching up to 180 m—well beyond the canopy height range considered in their experiments.
Similarly, \cite{ma2025} demonstrated that, although AlphaEarth Embeddings can achieve competitive performance in agricultural regression tasks, their spatial transferability remains limited.

\section{Conclusions}

This study investigated whether AlphaEarth Embeddings encode sufficient geospatial information to guide deep DL–based workflows for regional surface height mapping. 
The results showed that these embeddings can effectively support height inference when coupled with DL architectures. Both U-Net and U-Net++ substantially outperformed a linear Ridge Regression baseline, underscoring the importance of spatial context in decoding height-related information from embedding representations.

Among the evaluated models, U-Net++ ($R^2 = 0.84$, median difference = -2.62 m) demonstrated improved generalization compared to standard U-Net ($R^2 = 0.78$, median difference = -7.22 m), particularly under a shift in height frequency between the training and testing datasets. 
Robust statistics indicated reduced sensitivity to outliers and more stable error distributions for U-Net++, as reflected by smaller discrepancies between mean and median difference values and lower NMAD and RMSE on the testing set. These findings suggested that architectural design contributes to mitigating the effects of heterogeneous terrain conditions.

Importantly, while the testing results exhibited an elevated RMSE (approximately 16 m for U-Net++ and 19 m for U-Net) and a non-negligible bias, the models maintained a strong correlation with reference heights. 
This indicated that AlphaEarth Embeddings provide informative and transferable height-related patterns, even when applied to regions with differing height distributions. 
The observed performance degradation can be partially attributed to the substantial shift in height frequency between training and testing areas, highlighting the challenge of generalization under distribution mismatch rather than a lack of predictive height signal within the embeddings.
A temporal mismatch between the reference IGN DSM and the AlphaEarth Embeddings may also contribute to residual bias, as landscape changes occurring between acquisition periods (e.g., vegetation growth or infrastructure development) are not explicitly accounted for.
Despite these challenges, the results indicated that embedding-driven DL workflows represent a promising and computationally efficient approach for regional height mapping, particularly when lightweight convolutional architectures are employed as task-specific decoders.

Future work should expand the training dataset to increase spatial and morphological diversity, thereby improving robustness to height distribution shifts. 
Evaluating the framework across geographically distinct regions will be essential to rigorously assess transferability. 
In addition, benchmarking alternative embedding representations may provide further insight into how different pretraining strategies and dataset inclusion influence height inference performance and regional generalization capacity.

\section*{Declaration of competing interest}  

The authors declare that they have no known competing financial interests or personal relationships that could have appeared to influence the work reported in this paper.

\bibliographystyle{plainnat}
\bibliography{cit}
\newpage
\appendix
\refstepcounter{section}
\section*{\thesection\\Algorithm 1: Pseudocode of the UNET models}
\label{appendix1}
\hrule height 0.05cm
\:
{\footnotesize

\begin{algorithmic}[1]
\ttfamily
\State \textbf{Input:} AlphaEarth Embedding tensor $\mathtt{X} \in \mathbb{R}^{H \times W \times C_{in}}$, 
reference DEM $\mathtt{Y} \in \mathbb{R}^{H \times W \times C_{out}}$

\State \textbf{Parameters:} Patch size $\mathtt{P}=512$, batch size $\mathtt{B}=16$, 
learning rate $\mathtt{\eta}=1\times10^{-3}$, epochs $\mathtt{E}=500$

\State \textbf{Output:} Predicted DSM 
$\hat{\mathtt{Y}}_{\text{test}} \in \mathbb{R}^{H_{\text{test}} \times W_{\text{test}} \times C_{out}}$

\hrulefill

\State \textbf{\textcolor{blue}{Phase 1: Data Preparation}}
\State Acquire the embeddings and corresponding DSM:
\[
(\mathtt{X}_{\text{train}}, \mathtt{Y}_{\text{train}}),\;
(\mathtt{X}_{\text{test}}, \mathtt{Y}_{\text{test}})
\]

\State Construct patch dataset using patch size $\mathtt{P}$.
\State Split dataset into training (80\%) and validation (20\%).
\State Generate patches 
$\mathtt{X}_p \in \mathbb{R}^{P \times P \times C_{in}}$ and
$\mathtt{Y}_p \in \mathbb{R}^{P \times P \times C_{out}}$.

\hrulefill

\State \textbf{\textcolor{blue}{Phase 2: Model and Optimization Setup}}
\State Initialize model $\mathtt{M}$ as U-Net or U-Net++:
\[
\mathtt{M}=\texttt{Unet}(\texttt{encoder=ResNet18}, C_{in}=64, C_{out}=1)
\]

\State Initialize optimizer $\mathtt{O}$:
\[
\mathtt{O}=\texttt{AdamW}(\mathtt{M.parameters}, \eta, \lambda=10^{-4})
\]

\State Define loss function $\mathtt{L}$ as MSE.
\State Initialize ReduceLROnPlateau scheduler.
\State Set early stopping patience to 50 epochs.

\hrulefill

\State \textbf{\textcolor{blue}{Phase 3: Model Training}}
\For{$e=1$ to $\mathtt{E}$}
    \State Set model $\mathtt{M}$ to training mode
    \For{each batch $(\mathtt{X}_b,\mathtt{Y}_b)$}
        \State $\hat{\mathtt{Y}}_b \gets \mathtt{M}(\mathtt{X}_b)$
        \State Compute loss $\mathtt{L}(\hat{\mathtt{Y}}_b,\mathtt{Y}_b)$
        \State Backpropagate and update parameters
    \EndFor
    \State Evaluate validation loss
    \State Update learning rate
    \State Save model if validation loss improves
\EndFor

\hrulefill

\State \textbf{\textcolor{blue}{Phase 4: Inference and Evaluation}}
\State Predict DSM for test and train area:
\[
\hat{\mathtt{Y}}_{\text{test}}=\mathtt{M}(\mathtt{X}_{\text{test}})
\]

\State Compute evaluation metrics

\end{algorithmic}
}

\end{document}